%% file: main.tex
\documentclass[10pt,twocolumn,letterpaper]{article}

\usepackage{cvpr}              %

\input{preamble}
\definecolor{cvprblue}{rgb}{0.21,0.49,0.74}
\usepackage[pagebackref,breaklinks,colorlinks,citecolor=cvprblue]{hyperref}
\usepackage[symbol*]{footmisc} %

\def\paperID{1963}
\def\confName{CVPR}
\def\confYear{2024}

\title{NEAT: Distilling 3D Wireframes from Neural Attraction Fields}

\author{
    Nan Xue\textsuperscript{1} \quad
    Bin Tan\textsuperscript{1,2} \quad
    Yuxi Xiao\textsuperscript{1,3} \quad
    Liang Dong\textsuperscript{4} \quad
    Gui-Song Xia\textsuperscript{2} \quad
    Tianfu Wu\textsuperscript{5*} \quad 
    Yujun Shen\textsuperscript{1}
    \\[5pt]
    \textsuperscript{1}Ant Group \quad
    \textsuperscript{2}Wuhan University \quad
    \textsuperscript{3}Zhejiang University \quad
    \textsuperscript{4}Google Inc. \quad
    \textsuperscript{5}NC State University\\
}

\begin{document}
\twocolumn[{%
\renewcommand\twocolumn[1][]{#1}%
\maketitle
\vspace{-10mm}
    \begin{center}
    \begin{tabular}{c@{\hspace{1mm}}|c@{\hspace{1mm}}c@{\hspace{1mm}}c@{\hspace{1mm}}c@{\hspace{1mm}}c}
        \begin{tikzpicture}
            \node[anchor=south west,inner sep=0] (image) at (0,0) {\includegraphics[width=0.16\linewidth]{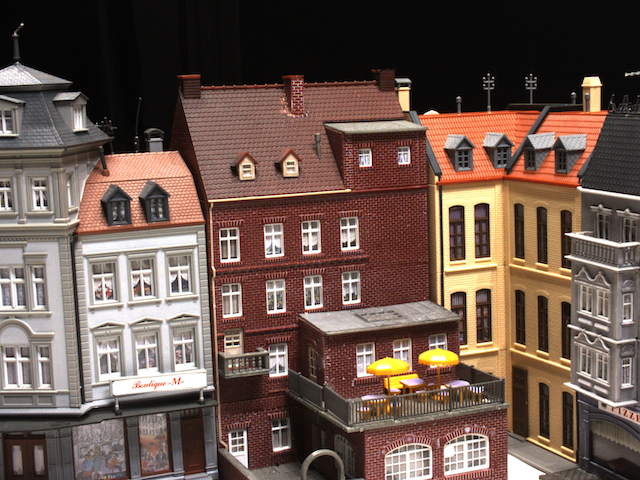}};
            \draw[black,fill=yellow] (image.south west) rectangle (image.south east) node[midway, yshift=-5pt, inner sep=5pt] {\scriptsize Image (DTU-23)};
        \end{tikzpicture}
        &
        \fbox{\includegraphics[width=0.16\linewidth]{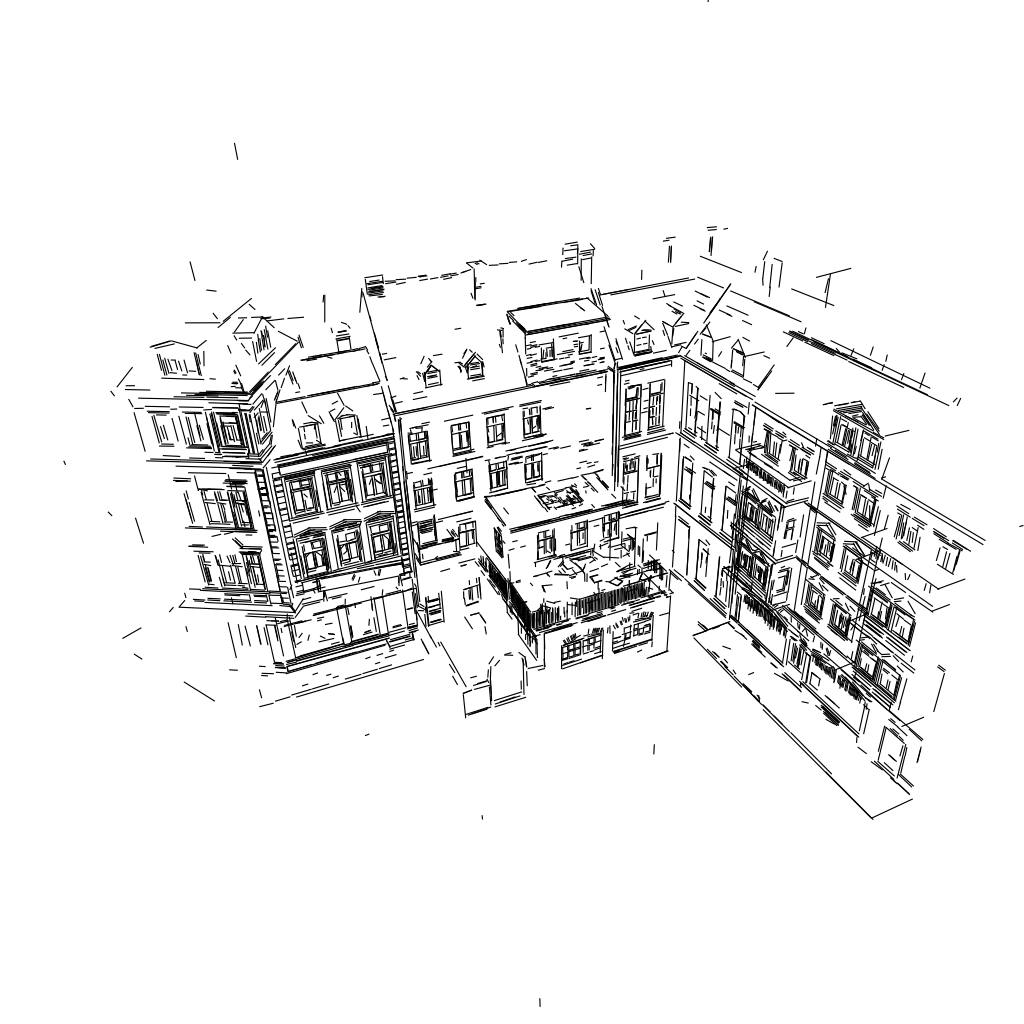}}
        \begin{tikzpicture}[overlay, remember picture,inner sep=0pt, outer sep=0pt]
            \node at (-1.7,0.2) [text=black] {\scriptsize Line3D++@LSD~\cite{HoferMB17-Line3D++}};
        \end{tikzpicture}&
        \fbox{\includegraphics[width=0.16\linewidth]{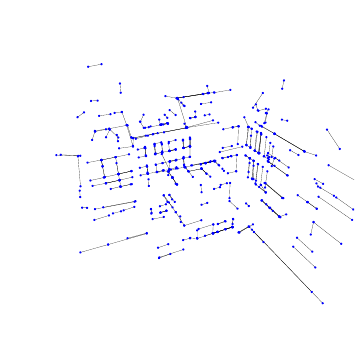}}
        \begin{tikzpicture}[overlay, remember picture,inner sep=0pt, outer sep=0pt]
            \node at (-1.7,0.2) [text=black] {\scriptsize Line3D++@HAWPv3~\cite{HoferMB17-Line3D++}};
        \end{tikzpicture}&
        \fbox{\includegraphics[width=0.16\linewidth]{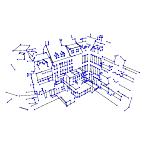}}
        \begin{tikzpicture}[overlay, remember picture,inner sep=0pt, outer sep=0pt]
            \node at (-1.7,0.2) [text=black] {\scriptsize LiMAP@HAWPv3~\cite{limap}};
        \end{tikzpicture}&
        \fbox{\includegraphics[width=0.16\linewidth]{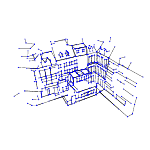}}
        \begin{tikzpicture}[overlay, remember picture,inner sep=0pt, outer sep=0pt]
            \node at (-1.7,0.2) [text=black] {\scriptsize NEAT (Ours)};
        \end{tikzpicture} \\

        \begin{tikzpicture}
            \node[anchor=south west,inner sep=0] (image) at (0,0) {\includegraphics[width=0.16\linewidth]{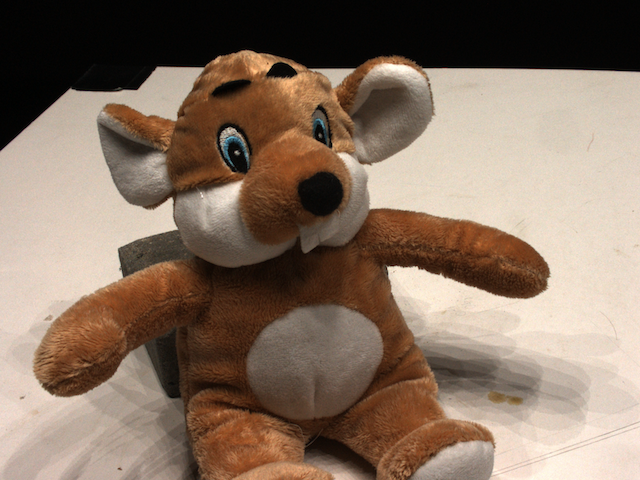}};
            \draw[black,fill=yellow] (image.south west) rectangle (image.south east) node[midway, yshift=-5pt, inner sep=5pt] {\scriptsize Image (DTU-105)};
        \end{tikzpicture}&
        \fbox{\includegraphics[width=0.16\linewidth]{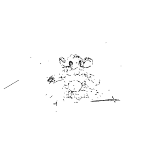}}
        \begin{tikzpicture}[overlay, remember picture,inner sep=0pt, outer sep=0pt]
            \node at (-1.7,0.2) [text=black] {\scriptsize Line3D++@LSD~\cite{HoferMB17-Line3D++}};
        \end{tikzpicture}&
        \fbox{\includegraphics[width=0.16\linewidth]{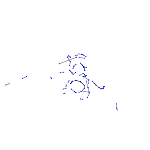}}
        \begin{tikzpicture}[overlay, remember picture,inner sep=0pt, outer sep=0pt]
            \node at (-1.7,0.2) [text=black] {\scriptsize Line3D++@HAWPv3~\cite{HoferMB17-Line3D++}};
        \end{tikzpicture}&
        \fbox{\includegraphics[width=0.16\linewidth]{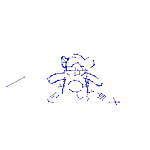}}
        \begin{tikzpicture}[overlay, remember picture,inner sep=0pt, outer sep=0pt]
            \node at (-1.7,0.2) [text=black] {\scriptsize LiMAP@HAWPv3~\cite{limap}};
        \end{tikzpicture}&
        \fbox{\includegraphics[width=0.16\linewidth]{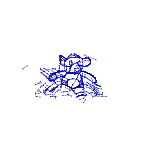}}
        \begin{tikzpicture}[overlay, remember picture,inner sep=0pt, outer sep=0pt]
            \node at (-1.7,0.2) [text=black] {\scriptsize NEAT (Ours)};
        \end{tikzpicture} 
    \end{tabular}
    \captionsetup{type=figure}
    \vspace{-2mm}
    \caption{%
        {{\bf Showcasing the evolution of 3D wireframe reconstruction}: The top reveals the transformative steps from a straight-line dominated urban landscape to an abstract wireframe, contrasting various methodologies. Below, the intricate transition from a curve-rich stuffed animal to its skeletal representation is depicted. While Line3D++~\cite{HoferMB17-Line3D++} and LiMAP~\cite{limap} utilize line-matching techniques, our novel NEAT approach forgoes matching, resulting in superior reconstruction fidelity with our proposed rendering-distilling formulation.
        }
    }
    \label{fig:teaser}
    \end{center}
    }]
\footnotetext[1]{Corresponding author.}

\input{sec/0_abstract}

\input{sec/1_intro}

\input{sec/2_relatedwork}

\input{sec/3_problem_state}

\input{sec/5_experiments}
\input{sec/6_conclusion}
{
    \small
    \bibliographystyle{ieeenat_fullname}
    \bibliography{ref}
}

\input{appx/0_overview}
\input{appx/1_video}

\input{appx/2_raysampling}
\input{appx/impl-details}

\input{appx/4_additional_exp_abc}
\input{appx/gaussian-rendering}

\input{appx/5_misc}

\end{document}

%% file: preamble.tex
\usepackage{capt-of}
\usepackage[dvipsnames]{xcolor}
\usepackage{xspace}
\usepackage{enumitem}
\usepackage{amsmath,amssymb,amsbsy,amsfonts,dsfont,pifont,bm,bbm,mathrsfs,mathtools,nicefrac}
\usepackage{algorithm,algpseudocode,listings}
\usepackage{booktabs,multirow,adjustbox,diagbox,threeparttable,tabularray}
\usepackage{tikz}
\usepackage{graphbox} %
\usepackage{wrapfig}
\usepackage{makecell}
\newcommand{\tocite}[1]{{\color{red} [TO CITE]}}

\definecolor{lightgray}{gray}{0.95} %
\usetikzlibrary{decorations.pathreplacing}

\newcommand{\myparagraph}[1]{\noindent\textbf{#1}}

%% file: sec/0_abstract.tex
\begin{abstract} \vspace{-4mm}
This paper studies the problem of structured 3D reconstruction using wireframes that consist of line segments and junctions, focusing on the computation of structured boundary geometries of scenes.
Instead of leveraging matching-based solutions from 2D wireframes (or line segments) for 3D wireframe reconstruction as done in prior arts, we present NEAT, a \textbf{rendering-distilling} formulation using neural fields to represent 3D line segments with 2D observations, and bipartite matching for perceiving and distilling of a sparse set of 3D global junctions. The proposed {NEAT} enjoys the joint optimization of the neural fields and the global junctions from scratch, using view-dependent 2D observations without precomputed cross-view feature matching. 
Comprehensive experiments on the DTU and BlendedMVS datasets demonstrate our NEAT's superiority over state-of-the-art alternatives for 3D wireframe reconstruction. 
Moreover, the distilled 3D global junctions by NEAT, are a better initialization than SfM points, for the recently-emerged 3D Gaussian Splatting for high-fidelity novel view synthesis using about 20 times fewer initial 3D points.
Project page: \url{https://xuenan.net/neat}.
\end{abstract}

%% file: sec/1_intro.tex
\section{Introduction}
\label{sec:intro}
\begin{figure}
    \centering
    \subfloat[\scriptsize Imagess \& 2D Wireframes\label{fig:input}]{
    \begin{minipage}{0.4\linewidth}
    \includegraphics[width=0.18\linewidth]{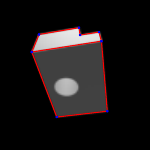}
    \includegraphics[width=0.18\linewidth]{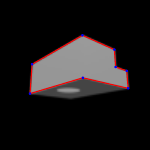}
    \includegraphics[width=0.18\linewidth]{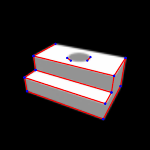}
    \includegraphics[width=0.18\linewidth]{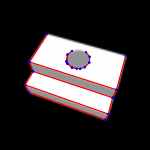}
    \includegraphics[width=0.18\linewidth]{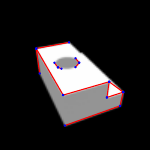}\\
    \includegraphics[width=0.18\linewidth]{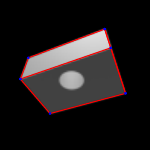}
    \includegraphics[width=0.18\linewidth]{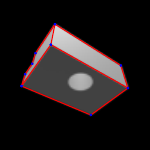}
    \includegraphics[width=0.18\linewidth]{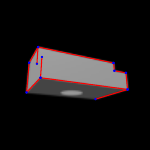}
    \includegraphics[width=0.18\linewidth]{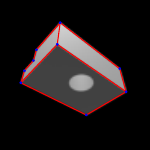}
    \includegraphics[width=0.18\linewidth]{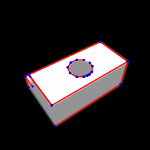}\\
    \includegraphics[width=0.18\linewidth]{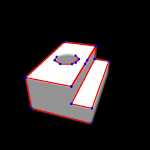}
    \includegraphics[width=0.18\linewidth]{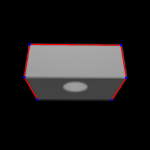}
    \includegraphics[width=0.18\linewidth]{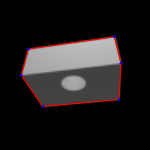}
    \includegraphics[width=0.18\linewidth]{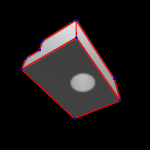}
    \includegraphics[width=0.18\linewidth]{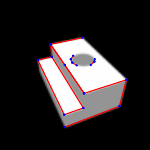}\\
    \includegraphics[width=0.18\linewidth]{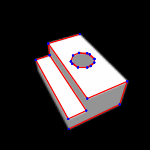}
    \includegraphics[width=0.18\linewidth]{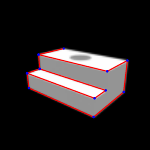}
    \includegraphics[width=0.18\linewidth]{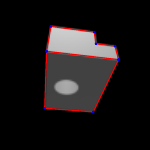}
    \includegraphics[width=0.18\linewidth]{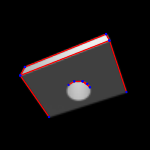}
    \includegraphics[width=0.18\linewidth]{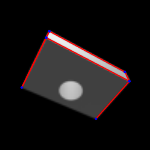}
    \end{minipage}
    }
    \subfloat[\scriptsize 3D Wireframe from Posed Images\label{fig:output}]{
    \begin{minipage}{0.5\linewidth}
    \fbox{\includegraphics[width=\linewidth]{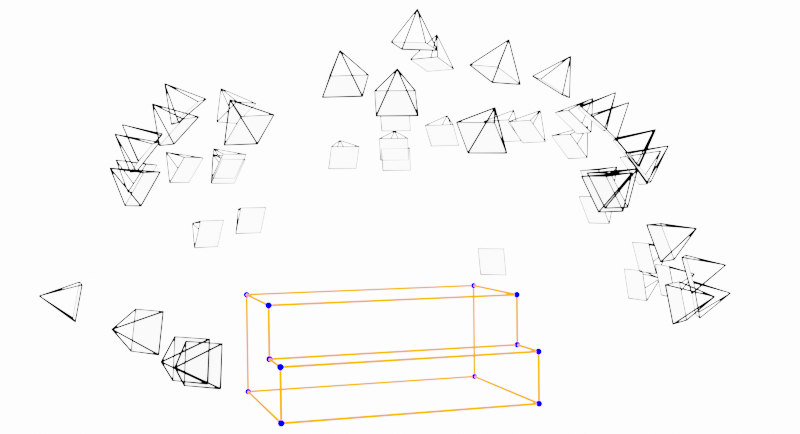}}
    \end{minipage}
    }
    \caption{{\bf Illustrative Overview} of the problem of 3D wireframe reconstruction. Given a set of posed images and the corresponding 2D wireframe detection results in \subref{fig:input}, the proposed NEAT estimates the 3D wireframe representation of the scene in \subref{fig:output}.
    }
    \label{fig:problem-figure}
\end{figure}

In this paper, we explore the field of multi-view 3D reconstruction, drawing inspiration from the paradgim of the primal sketch proposed by D. Marr~\cite{marr2010vision}. 
Our objective is to develop a concise yet precise representation of 3D scenes, derived from multi-view images with known camera poses. Specifically, our focus is on wireframe representations~\cite{Zhou-Manhattan-Wireframes,lcnn,hawp,hawpv3}, which define the boundary geometry of scene images through line segments and junctions as the 2D wireframe representation. We dedicate our efforts to advancing the reconstruction of 3D wireframes based on their 2D counterparts detected in multi-view images, as shown in \cref{fig:teaser} and \cref{fig:problem-figure}.

The challenge of \emph{multi-view 3D wireframe reconstruction} has been previously explored within the realm of line-based 3D reconstruction~\cite{HoferMB17-Line3D++,ELSR,limap}, primarily following the feature triangulation pipelines~\cite{colmapCVPR16}, which heavily rely on the accuracy of multi-view feature correspondences. Various methods have been developed to enhance this accuracy~\cite{ELSR,gluestick,sold2}. However, a significant challenge arises from view-dependent occlusions of line features: when projecting a 3D line segment onto 2D images, the endpoints of the line segment may be truncated in the 2D projections by chance. Such discrepancies can severely impact the accuracy of the reconstruction, as the matching process relies on these endpoints to accurately represent the 3D geometry. These matching-based methods often result in incomplete 3D line models or suffer from fragmentation and noise, depending on the choice of 2D detectors~\cite{von-LSD,deeplsd,hawp,hawpv3,afm-cvpr,afm-pami} and matchers~\cite{sold2,gluestick} of line segments, as in \cref{fig:teaser}.

\myparagraph{Dense Fields of Sparse Geometries.} We challenge the explicit matching pipeline of 3D wireframe reconstruction from the perspective of dense field representation.
We draw inspiration from the ``{\bf implicit matching}" capacity~\cite{levels2fm} of the emerging neural implicit fields~\cite{Mip-NeRF,YarivGKL21-VolSDF,DeepSDF} for 3D dense representations (\eg, density fields and signed distance functions), and propose to {\em render} 3D line segments from multi-view 2D observations. 
Such a basic idea roughly works by leveraging a coordinate MLP to render 3D line segments from 2D observations, but remains problematic due to the entailed view-by-view rendering of 3D line segments in two-fold: (1) the 2D line segments of a detected wireframe often undergo localization errors, resulting in erroneous 3D line segment predictions via view-by-view rendering, and (2) simply stacking the rendered 3D line segments from all views leads to a very large amount of 3D line segments, requiring non-trivial merging/fusion to form a 3D wireframe representation of the scene. 

\myparagraph{Line-to-Point Attraction in Neural Fields.} We tackle the above issues by leveraging the line-to-point attraction that inherently persists in the wireframe representation, in which {\em every endpoint of a 3D line segment should be in the set of 3D junctions of the underlying scene}. 
Based on this, we formulate the two types of entities of 3D wireframes, the 3D line segments and junctions, in a novel {\em rendering-distilling} formulation, where the sparse set of 3D line segments are represented in a dense neural field while the junctions play the role of distilling a sparse wireframe structure from the fields. Our work is entitled as {\em NEural Attraction} (NEAT) for 3D wireframe reconstruction, mainly because of the neural design of the 3D line segments and junctions, and of leveraging the line-to-point attraction to enable joint optimization of the neural networks from multi-view images and its 2D wireframe detection results. To the best of our knowledge, we accomplish the first matching-free solution of 3D wireframe/line reconstruction by learning and optimizing from random initializations without any 3D scene information required.

In experiments, we showcase that our matching-free NEAT solution significantly outperforms all the matching-based approaches with accurate yet complete 3D wireframe reconstruction results on both the DTU~\cite{DTU-AanaesJVTD16} and BlendedMVS~\cite{BMVS-dataset} datasets, working well in both straight-line dominated scenes and curve-based (or polygonal line segment dominated) scenes that challenges the traditional matching-based approaches, paving a way towards learning 3D primal sketch in a more general way. Furthermore, we show that the neurally perceived 3D junctions is applicable to the recently proposed 3D Gaussian Splatting~\cite{kerbl3Dgaussians} as better initialization than the COLMAP~\cite{colmapCVPR16} with about 20 times fewer points, showing case the potential of structured and compact 3D reconstruction.

%% file: sec/2_relatedwork.tex
\section{Related Work}
\myparagraph{Structured 3D Reconstruction in Geometric Primitives.}
Because of the inherent structural regularities for scene representation conveyed by line structures~\cite{marr2010vision,AutoLineMatch,GeoRecovery,DecompositionScene,PolyhedralScene} and planar structures~\cite{planetr,nopesac}, there has been a vast body of literature on line-based multiview 3D reconstruction tasks including single-view 3D reconstruction~\cite{how3d,planetr}, line-based SfM~\cite{SalaunMM16a,ChandrakerLK09}, SLAM~\cite{PLSLAM,Line-Flow-SLAM}, and multi-view stereo~\cite{HoferMB17-Line3D++,limap,ELSR} based on the theory of multi-view geometry~\cite{hartley-zisserman-mvg}. Due to the challenge of line segment detection and matching in 2D images, most of those studies expected the 2D line segments detected from input images to be redundant and small-length to maximize the possibility of line segment matching. As for the estimation of scene geometry and camera poses, the keypoint correspondences (even including the 3D point clouds) are usually required. 
For example in Line3D++~\cite{HoferMB17-Line3D++}, given the known camera poses by keypoint-based SfM systems~\cite{colmapCVPR16,colmapECCV16,Theia,visualSfM}, it is still challenging though to establish reliable correspondences for the pursuit of structural regularity for 3D line reconstruction. For our goal of 3D wireframe reconstruction, because 2D wireframe parsers aim at producing parsimonious representations with a small number of 2D junctions and long-length line segments, those correspondence-based solutions pose a challenging scenario for cross-view wireframe matching, thus leading to inferior results than the ones using redundant and small-length 2D line segments detected by the LSD~\cite{von-LSD}. To this end, we present a correspondence-free formulation based on coordinate MLPs, which provides a novel perspective to accomplish the goal of 3D wireframe reconstruction from the parsed 2D wireframes.

\myparagraph{Neural Rendering for Geometric Primitives.} In recent years, the emergence of neural implicit representations~\cite{Ben-NeRF,Mip-NeRF,IDR,IshitMehta-LevelSet} have greatly renown the 3D vision community. By using coordinate MLPs to implicitly learn the scene geometry from multi-view inputs without knowing either the cross-view correspondences or the 3D priors, it has largely facilitated many 3D vision tasks including novel view synthesis, multi-view stereo, surface reconstruction, \etc. Some recent studies further exploited the neural implicit representations by (explicitly and implicitly) taking the geometric primitives such as 2D segmentation masks into account to lift the 2D detection results into 3D space for scene understanding and interpretation~\cite{Panoptic-NeRF,Panoptic-Neural-Fields,DM-NeRF,Object-SDF}. Most recently, nerf2nerf~\cite{nerf2nerf} exploited a geometric 3D representation, surface fields as a drop-in replacement for point clouds and polygonal meshes, and takes the keypoint correspondences to register two NeRF MLPs. Our study can be categorized as the exploration of geometric primitives in neural implicit representation, but we focus on computing a parsimonious representation by using the most fundamental geometric primitives, the junction (points) and line segments, to provide a compact and explicit representation from coordinate MLPs.

%% file: sec/3_problem_state.tex
\section{NEAT of 3D Wireframe Reconstruction}
In this section, we formulate the problem of 3D wireframe reconstruction, lying on the high-level idea of approaching the goal of 
using volume rendering instead of the explicit line segment matching to build a unified 3D computational representation of line segments and junctions from the 2D detected wireframes. 

\myparagraph{Problem Statement.} 
For the problem illustrated in \cref{fig:problem-figure}, we present our approach for 3D wireframe reconstruction from $n$-view posed images, $\{\mathcal{I}_i\}_{i=1}^n$. Each image $\mathcal{I}_i$ is characterized by intrinsic and extrinsic matrices. We use the HAWPv3 model~\cite{hawpv3} to detect 2D wireframes in these images, represented as undirected graphs ${G_i = (V_i, E_i)}$. The goal is to construct a 3D wireframe graph $\mathcal{G} = (\mathcal{V}, \mathcal{E})$, translating these 2D wireframes into a 3D representation with $\mathcal{V}$ as 3D junctions and $\mathcal{E}$ as the 3D line segments.

\begin{figure}
    \centering
    \includegraphics[width=0.95\linewidth]{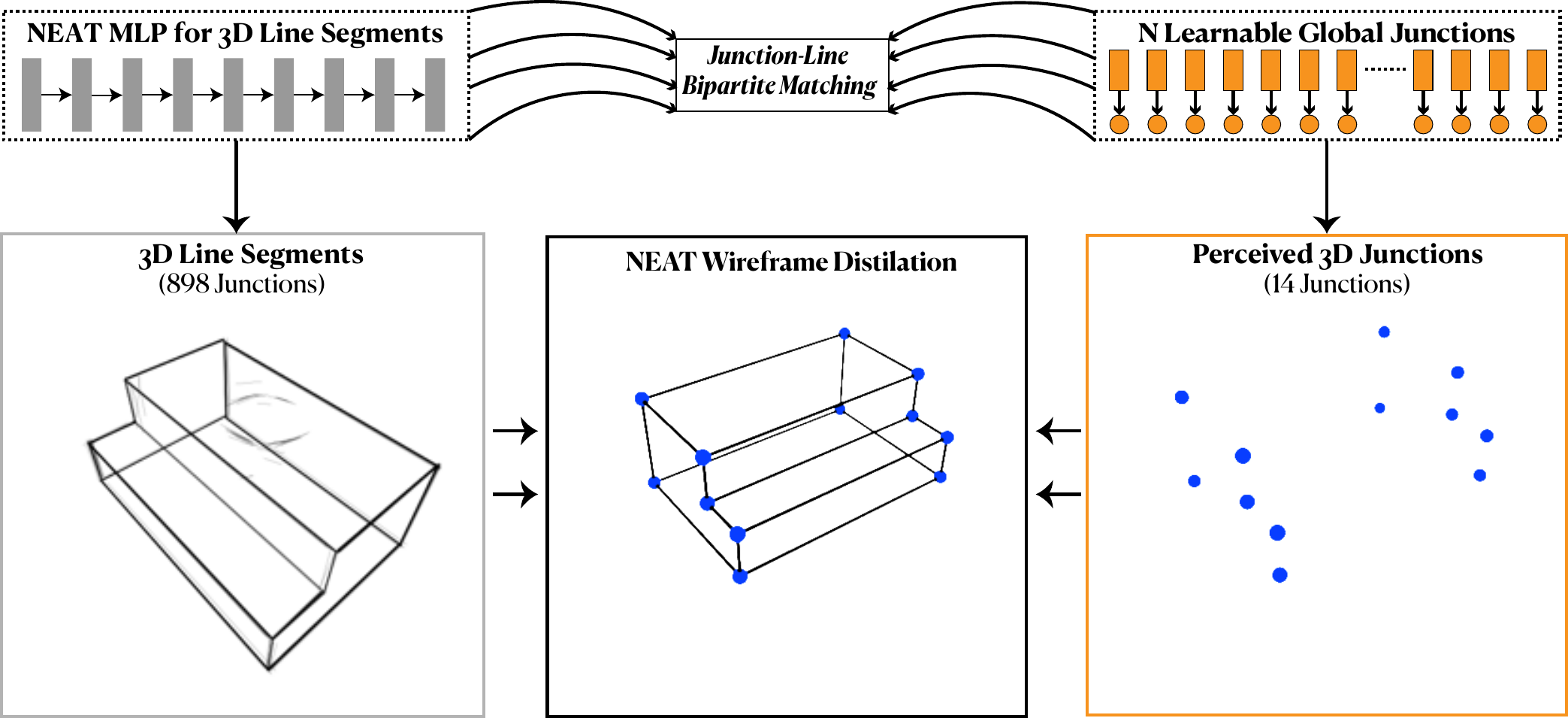}

    \vspace{0.5em}
    \caption{The proposed NEAT field learning framework for 3D wireframe reconstruction. In the top, the neural design of NEAT MLP and the predefined $N$ global junctions are illustrated, these two components are ``attracted" by the junction-to-line bipartite matching, resulting a rendering-yet-distillation formulation to render 3D line segments in NEAT MLP as a dense representation of 3D line segments, and then distilled by the learned 3D global junctions for wireframe reconstruction.
    }
    \label{fig:neat-illustration}
\end{figure}

\myparagraph{Method Overview.} 
Our NEAT method is built on the VolSDF framework~\cite{YarivGKL21-VolSDF} with two primary neural components: (1) a Neural Attraction Field for 3D line segments, and (2) a Global 3D Junction Perceiver (GJP). These components work jointly to create NEAT 3D wireframe models from the 2D wireframe observations. We start by learning a dense representation of 3D line segments from 2D wireframes using the Neural Attraction Field, as visualized in Figure \ref{fig:neat-illustration}. This is followed by the Global 3D Junction Perceiver, which identifies a set of 3D junctions. As a final step of the wireframe reconstruction, the perceived 3D junctions play in a distillation role to clean up the optimized NEAT field. 
In implementation, we adopt a simple design for the MLPs used in the SDF and radiance field, aligned with VolSDF specifications. For the NEAT field, a 4-layer MLP renders the 3D line segments. Additional implementation details and hyperparameters are outlined in the \cref{appx:details-neat}. %

\subsection{Rendering 3D Line Segments from 2D}\label{sec:line-render}
We propose to leverage the power of ``implicit matching" ability of neural 
fields to obtain 3D line segments. Our method is built on the basic formulation of VolSDF~~\cite{YarivGKL21-VolSDF} that renders a ray $\mathbf{x}_t = \mathbf{c} + t\cdot \mathbf{v}$ emanating from the camera location $\mathbf{c}\in\mathbb{R}^3$ with the (unit) view direction $v \in \mathbb{R}^3$ to estimate the image appearance by,
\begin{equation}\label{eq:expectedcolor}
    \hat{I}(c, v) = \int_{0}^{\infty} T(t)\cdot \sigma(x_t) \cdot \mathbf{r}(x_t, v, \mathbf{n}(x_t), z(x_t)) dt, 
\end{equation}
where $\mathbf{r}(\cdot)$ is the radiance of the ray $x_t$, and $T(t)$ is the transmittance $T(t) = \exp -\int_{0}^t \sigma\left(x(s)\right)ds$ along the ray from camera center to $t$, the density field $\sigma(\cdot)$ is transformed by the signed distance function $d_{\Omega}(\mathbf{x})$ of an implicit field using, 
\begin{equation}\label{eq:sdf2density}
    \sigma(\mathbf{x}) = \frac{1}{\beta} \Psi_{\beta}(-d_{\Omega}(\mathbf{x})),
\end{equation}
with the learnable scaling factor $\beta$. As for the optimization of SDF and radiance fields, the image loss $\mathcal{L}_{\rm img}$ between the rendered image $\hat{I}$ and its corresponding ground-truth $\mathcal{I}$, and Eikonal loss $\mathcal{L}_{\rm eik}$ for SDF network are used.

\paragraph{Neural Attraction Fields.} 
In our NEAT method, we adapt volume rendering, typically used for optimizing dense 3D representations like density fields and SDFs, to focus on 3D line segments and junctions. Our approach is inspired by the dense attraction field representations used in 2D line segment detection and wireframe parsing, as extensively researched in previous studies~\cite{hawp,hawpv3}. As illustrated in \cref{fig:neat-illustration} using a synthetic example, we utilize the attracted pixels of 2D line segments in each image to define the rays for 3D rendering. For each segment, its attracted pixels are projected perpendicularly onto the 2D segment. This projection is confined within the endpoints of the segment with respect to a predefined distance threshold, $\tau_{\rm ray}$. Each pixel is associated with its nearest line segment, ensuring a dense coverage of supporting areas for the segments. This approach facilitates the volume rendering of 3D line segments by providing a robust underlying structure.

In our approach, we model a 3D line segment at any point $\mathbf{x}_t$ along a ray. The endpoint displacements $(\Delta \mathbf{x}_t^1, \Delta \mathbf{x}_t^2)$ relative to $\mathbf{x}_t$ are computed as,
\begin{equation}
(\Delta \mathbf{x}_t^1, \Delta \mathbf{x}_t^2) = L(\mathbf{x}_t) \in \mathbb{R}^{2\times 3},
\end{equation}
yielding the two endpoints of the segment by $(\mathbf{x}_t + \Delta \mathbf{x}_t^1, \mathbf{x}_t + \Delta \mathbf{x}_t^2)$. The mapping function $L(\cdot)$ is parameterized by a 4-layer coordinate MLP. It incorporates the view direction $\mathbf{v}$, the surface normal $\mathbf{n}(\cdot)$ from the SDF gradient, and a 128-dimensional feature vector $\mathbf{z}(\mathbf{x}_t)$ from the SDF network, reflecting the view-dependent nature of 2D line segments.
For rendering a 3D line segment, we apply the equation,
\begin{equation}
(\mathbf{x}^s,\mathbf{x}^t) = \int_{0}^{\infty} T(t)\sigma(t) \left(L(\mathbf{x}_t) + \mathbf{x}_t\right) dt.
\end{equation}
Here, $\mathbf{x}^s$ and $\mathbf{x}^t$ are the 3D endpoints for the attraction pixel $\mathbf{x}$ of a 2D line segment $\ddot{l} = (\jmath_1, \jmath_2) \in V_i\times V_i$ of the $i$-th view, calculated along its ray $\mathbf{x}_t$.

According to the pixel-to-line relationship defined by 2D attraction field representations, the rendered 3D line segment $(\mathbf{x}^s,\mathbf{x}^t)$ of a ray $\mathbf{x}_t$ should be consistent with $\ddot{l} = (\jmath_1,\jmath_2)$,
thus resulting in a loss function between the projected 2D endpoints by viewpoint projection $\Pi(\cdot)$ and $\ddot{l}$ in, 
\begin{equation}
    \mathcal{L}_{\rm neat} = \left\|\Pi(\mathbf{x}^s)-\jmath_1\right\|^2 + \left\|\Pi(\mathbf{x}^t)-\jmath_2\right\|^2.
\end{equation}
The proposed Neural Attraction Fields of 3D line segments is optimized together with SDF and the radiance field by minimizing the loss functions stated above, forming a querable and dense representation of 3D line segments. %

\begin{figure}
    \centering
    \subfloat[898 Lines (100 Views)]{
    \fbox{\includegraphics[width=0.4\linewidth]{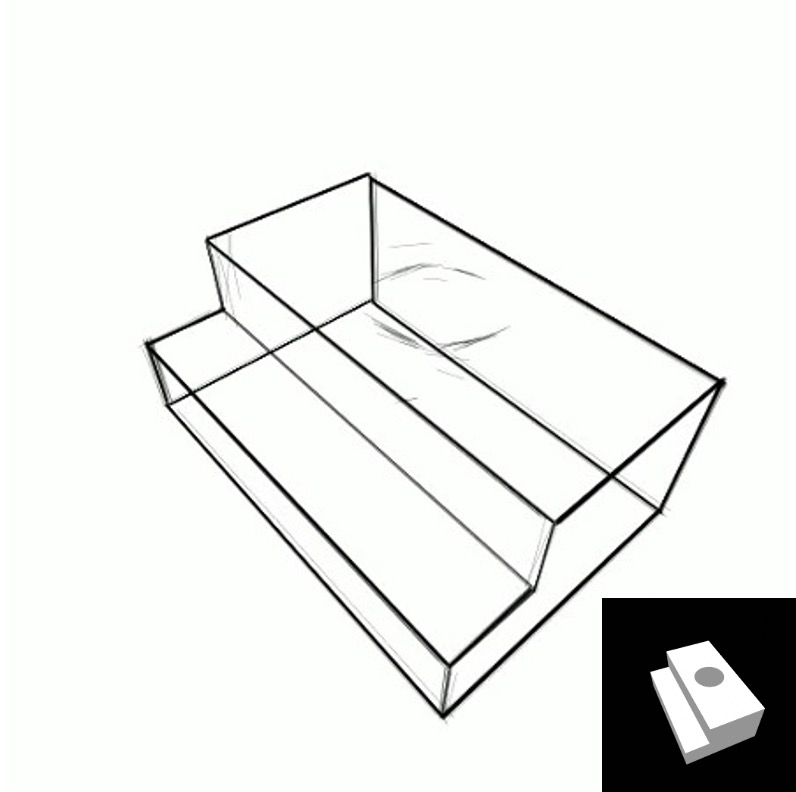}}
    }
    \subfloat[6731 Lines (49 Views)]{
    \fbox{\includegraphics[width=0.4\linewidth]{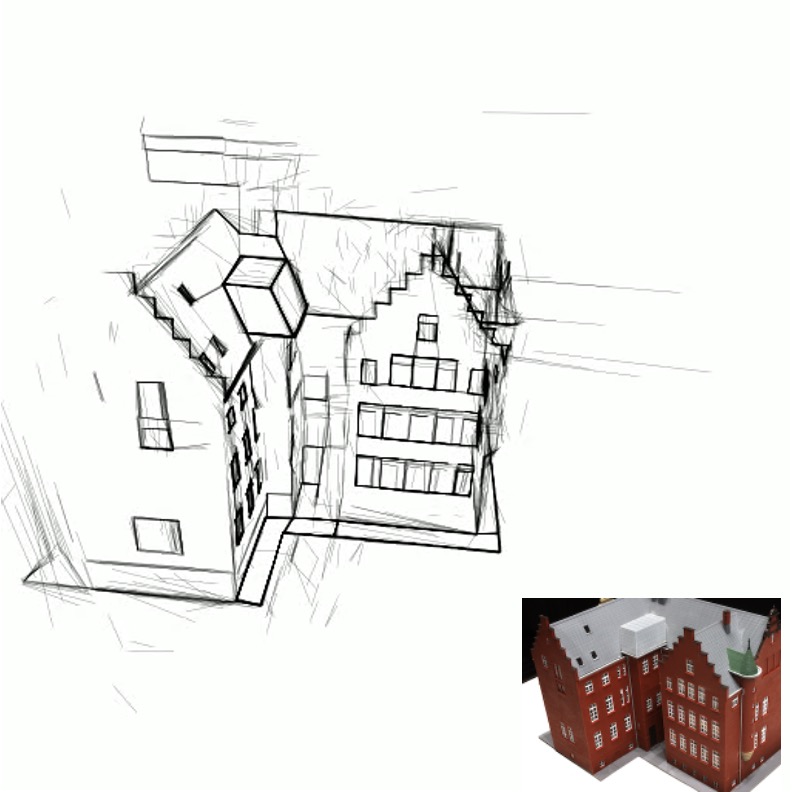}
    }
    }
    \caption{Two cases of learned noisy and redundant 3D line segments by line segment rendering. The case (a) takes the images and line segments introduced in \cref{fig:input}, and the case (b) is a real-world case of DTU-24 scene.}
    \label{fig:noisy-lcd}
\end{figure}

Minimizing the loss functions $\mathcal{L}_{\rm neat}$, $\mathcal{L}_{\rm img}$, and $\mathcal{L}_{\rm eik}$ allows us to derive a geometrically meaningful but noisy 3D line cloud from multi-view images, as demonstrated in \cref{fig:noisy-lcd} using both a synthetic example and a real case from the DTU-24 scene~\cite{DTU-AanaesJVTD16}. The absence of explicit line matching across multiple views leads to duplication of the same 3D line segments, each with its own view-dependent prediction errors. In the following section, we discuss how this redundancy and noise, while initially seeming detrimental, actually provide a strong inductive bias towards achieving the goal of 3D wireframe reconstruction.

\subsection{Neural 3D Junction Perceiver} \label{sec:junctions}
This section introduces our method to ``clean up" the noisy and redundant 3D line cloud created by Neural Attraction Fields. Leveraging the relationship between 3D junctions and line segments in wireframes, we propose a neural and joint optimization approach, central to our NEAT method. Using the 3D line cloud, denoted by $\mathbf{L}_{\rm neat}$, a query-based learning method is designed for perceiving 3D junctions  (\cref{eq:junctions}) via junction-line attraction, which plays the role of distillation for 3D wireframe reconstruction.

\myparagraph{Global 3D Junction Percieving.} Our 3D line segment rendering inherits the dense representation as the density field and the radiance field. To achieve parsimonious wireframes,  we propose a novel query-based design to holistically perceive a predefined sparse set of $N$ 3D junctions by 
    \begin{equation}
        Q_{N\times C}  \xrightarrow[]{\text{MLP}} J_{N\times 3}, \label{eq:junctions}
    \end{equation}
    where $Q_{N\times C}$ are $C$-dim latent queries (randomly initialized in learning). Surprisingly, as we shall show in experiments, the underlying 3D scene geometry induced synergies between $J_{N\times 3}$ and the above 3D line segment rendering integral enable us to learn a very meaningful global 3D junction perceiver. %

In the absence of well-defined ground-truth for learning 3D junctions, we use the endpoints of redundant rendered 3D line segments (\cref{sec:line-render}) as noisy labels. By reshaping the line cloud $\mathbf{L}_{\rm neat}$ into $\mathbf{J}_{\rm neat}\in \mathbb{R}^{2M\times 3}$, our process involves two steps: (1) clustering $\mathbf{J}_{2M\times 3}$ using DBScan to yield pseudo 3D junctions $\mathbf{J}_{\rm cls}\in\mathbb{R}^{m\times 3}$ with $m<2M$ clusters; (2) applying bipartite set-to-set matching between the perceived junctions $J_{N\times 3}$ (\cref{eq:junctions}) and $\mathbf{J}_{\rm cls}$ using the Hungarian algorithm. The matching cost is based on the $\ell_2$ norm between 3D points. We define $\mathcal{J}=\{(J_{k}, \mathbf{J}_{i_k}^{\rm cls})|k = 1,\ldots, K\}$ as the set of matched junctions, where $K=\min(N, m)$, and $i_k$ is the index of the $k$-th matched pseudo label $\mathbf{J}_{i_k}^{\rm cls}$. 
Then, our goal is to minimize the distance between matched junctions and their corresponding pseudo labels using 
\begin{equation}
    \mathcal{L}_{jc}(J_k, \mathbf{J}_k) = \left \|J_k - \mathbf{J}_k \right\|_1 + \lambda\cdot \left \|\Pi(J_k) - \Pi(\mathbf{J}_k) \right\|_1,
\end{equation}
where $\Pi()$ is the 3D-to-2D projection, and $\lambda$ the trade-off parameters (e.g. 0.01 in our experiments). 

\begin{figure}
    \centering
    \resizebox{0.8\linewidth}{!}{
    \begin{tabular}{|c|c|c|}
    \hline
    \includegraphics[width=0.3\linewidth]{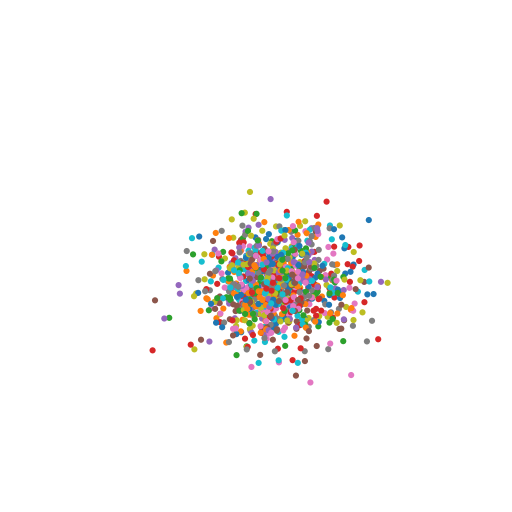} & 
    \includegraphics[width=0.3\linewidth]{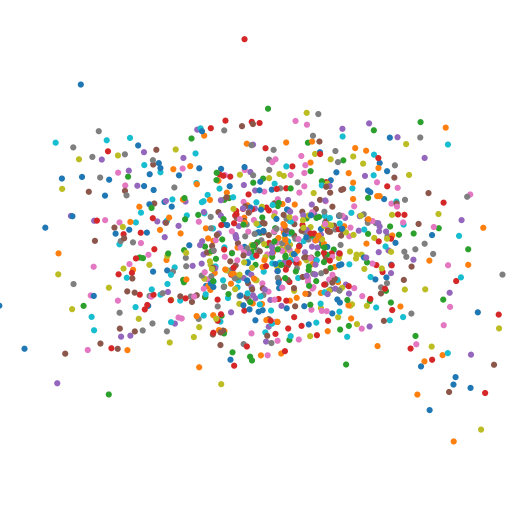} & 
    \includegraphics[width=0.3\linewidth]{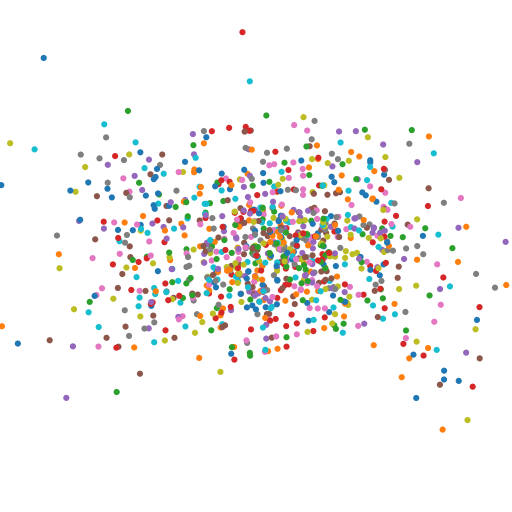} \\\hline
    \includegraphics[width=0.3\linewidth]{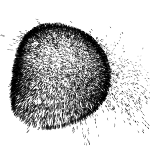}&
    \includegraphics[width=0.3\linewidth]{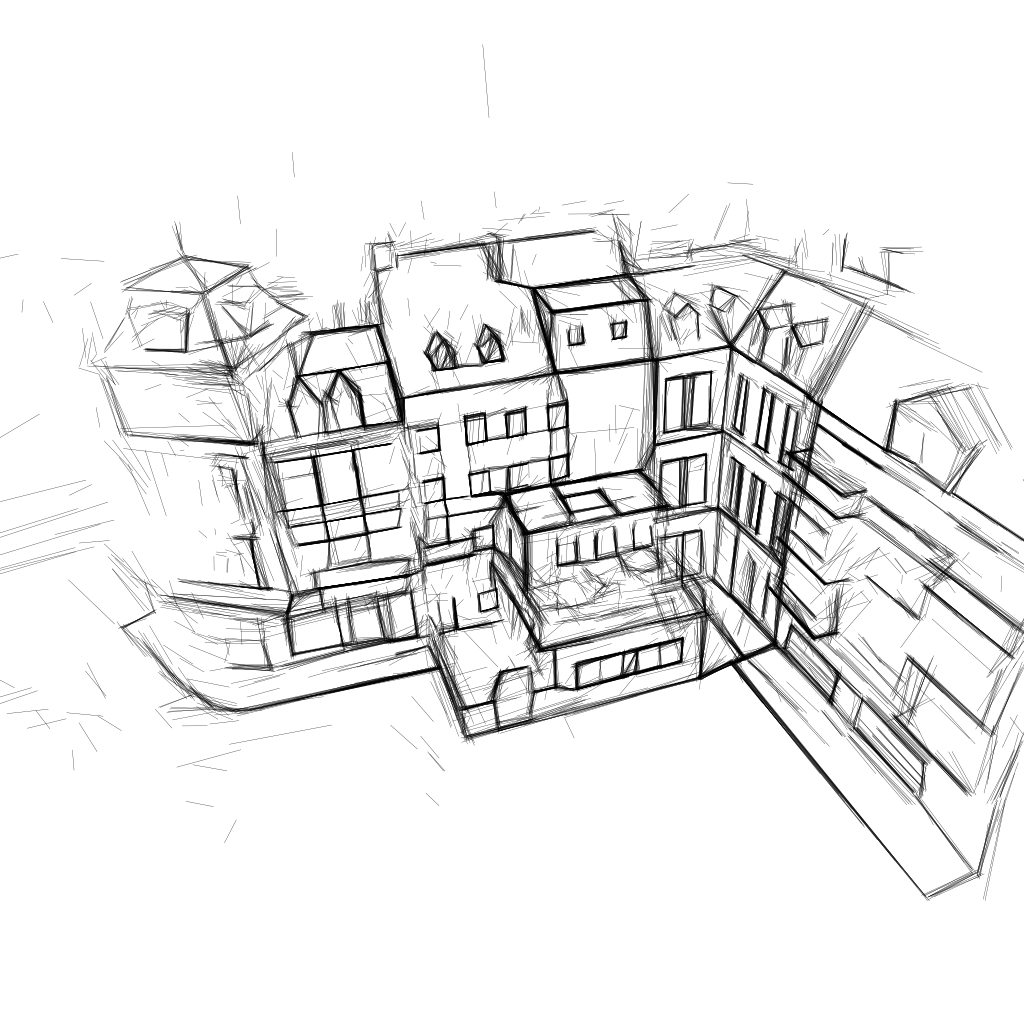}&
    \includegraphics[width=0.3\linewidth]{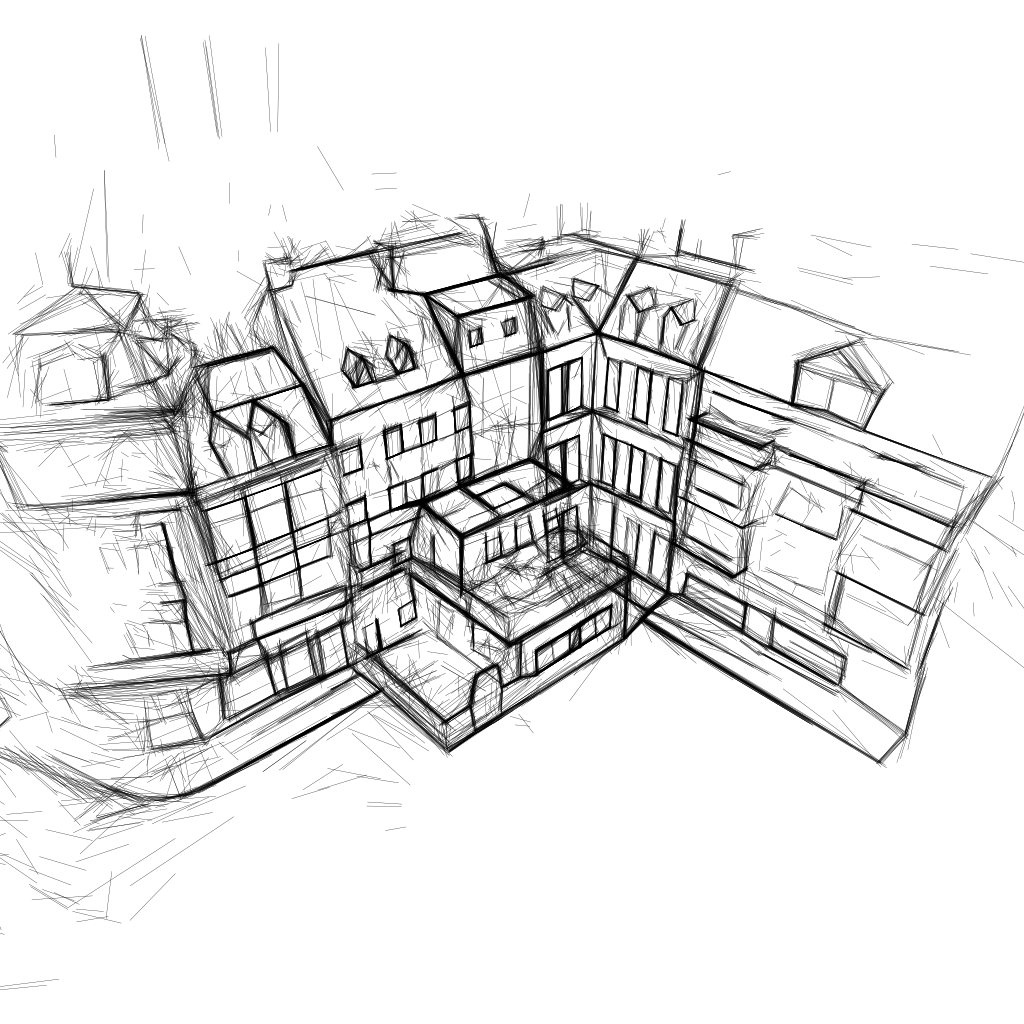} \\
    \hline
    \multicolumn{1}{c}{(a). Random Inits.} & 
    \multicolumn{1}{c}{(b). 24K iterations} & 
    \multicolumn{1}{c}{(c). Final iteration} 
    \end{tabular}
    }
    \caption{{\bf Optimization Process} of 3D Junction Perceiving (top) from the noisy 3D line cloud (bottom) on the DTU-23 scene.
    }
    \label{fig:junctions-it}
\end{figure}

\myparagraph{Joint Optimization.} 
In our final implementation, we refine our approach by jointly optimizing the NEAT field and the 3D junction perceiver. This optimization involves minimizing all aforementioned loss functions in a weighted sum, which allows for dynamic distillation of 3D junctions from the noisy 3D line cloud generated by the NEAT field. The total loss function, $\mathcal{L}_{\rm total}$, is expressed as:
\begin{equation}
    \mathcal{L}_{\rm total} = \mathcal{L}_{\rm img} + \lambda_e \mathcal{L}_{\rm eik} + \lambda_n \mathcal{L}_{\rm neat} + \lambda_j \mathcal{L}_{jc},
\end{equation}
where $\mathcal{L}_{\rm img}$ and $\mathcal{L}_{\rm eik}$ are as defined in \cite{YarivGKL21-VolSDF}. The weights $\lambda_n, \lambda_e,$ and $\lambda_j$ are all set to 0.01. As depicted in Figure \ref{fig:junctions-it}, this optimization process continually refines the global 3D junctions by extracting them from the 3D line cloud of NEAT field at each iteration, all  trained from scratch.

\subsection{NEAT Wireframe Distillation using Junctions}
After training, we acquire $N$ 3D junctions $J_{N\times 3}$ and $M$ 3D line segments $\mathbf{L}_{\rm neat} \in \mathbb{R}^{M\times 2\times 3}$. The line segments are indexed by 3D junctions based on their spatial relationship, assigning each segment $\mathbf{L}_{\rm neat}^{i}$ a global ID within $(u,v) \in \{0,\ldots,N-1\}\times \{0,\ldots,N-1\}$, with $u<v$. Indexing is informed by endpoint distances. Segments with angular distances over 10 degrees or perpendicular distances above 0.01 units in 3D space are deemed "too far" and removed, ensuring alignment with the 3D junctions. Further details are available in \cref{appx:ray-sampling}. 

Endpoint indexing significantly reduces the number of 3D line segments. Segments like $(\mathbf{x}^s_i, \mathbf{x}^t_i)$ and $(\mathbf{x}^s_j, \mathbf{x}^t_j)$ sharing the same junction IDs $(u_i,v_i) = (u_j,v_j) = (u,v)$ are grouped under one global line segment defined by $(u,v)$. We represent these grouped segments as $\mathbf{L}_{u,v} = \{\mathbf{l}_{u,v}^{1},\ldots, \mathbf{l}_{u,v}^{T}\}\in \mathbb{R}^{T\times 2\times 3}$, where $T=T_{u,v}$ indicates the count of segments in $\mathbf{L}_{u,v}$. For convenience, the global line segment for index $(u,v)$ is denoted as $\mathbf{l}_{u,v}^0 = (J_u,J_v)$. Junctions not indexed by more than one line segment in $\mathbf{L}_{\rm neat}$ are marked as inactive.

\myparagraph{The 3D Wireframe.} 
After indexing the 3D line segments $\mathbf{L}_{\rm neat}$ with global junctions, we form the graph $\mathcal{G} = (\mathcal{V}, \mathcal{E})$ composed of active global junctions and their index pairs. To refine this graph, we remove isolated junctions and line segments, resulting in the final 3D wireframe $\mathcal{G}$, where $\mathcal{V} \subset \mathbb{R}^3$ represents the vertices and $\mathcal{E} \subset \mathbb{Z}^2$ the edges.

\myparagraph{Least Square Optimization of 3D Junctions.} 
Given that 3D junctions are derived from a noisy 3D line cloud, we optimize them by leveraging their relationships with global line segments $(J_u,J_v)$ and corresponding 3D line segments $\mathbf{L}(u,v)$. This alignment aims to match junctions with their supporting 3D line segments. The optimization is framed as a non-linear least squares problem with the cost function $\mathcal{L}(J)$, defined as:
\begin{equation}
   \mathcal{L}(J) = \sum_{(u,v)} \sum_{i=1}^{T_{u,v}} d_{\rm ang}(\mathbf{l}_{u,v}^0, \mathbf{l}_{u,v}^{i})^2 + d_{\rm perp}(\mathbf{l}_{u,v}^0, \mathbf{l}_{u,v}^{i})^2,
\end{equation}
where $d_{\rm ang}$ and $d_{\rm perp}$ represent the angular and perpendicular distances between two 3D line segments, respectively. The optimization details are provided in \cref{appx:finalization}.

\myparagraph{The Final Wireframe.} After leveraging the least square optimization to adjust the position 3D junctions, we further remove the isolated junctions and the isolated line segments in $\mathcal{G}$ of which their projection to 2D space are not supported by any line segment of the 2D wireframe observations. Here, the criterion of the support is defined by the minimum angular distance and the perpendicular distance between the projected 3D line segment and the 2D line segment is not more than $10$ degree and $5$ pixels, respectively. After the filtering, we adjust the actived 3D junctions by querying SDF, see \cref{appx:finalization} for details.

%% file: sec/5_experiments.tex
\section{Experiments}

In experiments, we mainly testify our NEAT on two datasets (\ie, the DTU dataset~\cite{DTU-AanaesJVTD16} and the BMVS dataset~\cite{BMVS-dataset}) for real-scene multiview images with known camera poses. In addition to those two datasets, in \cref{appx:exp-abc}, the experiments on the ABC dataset~\cite{abc-dataset} evaluated by using the 3D wireframe annotations further verified our proposed NEAT approach for the 3D wireframe representation.
\begin{figure*}
    \centering
     \includegraphics[width=0.97\linewidth]{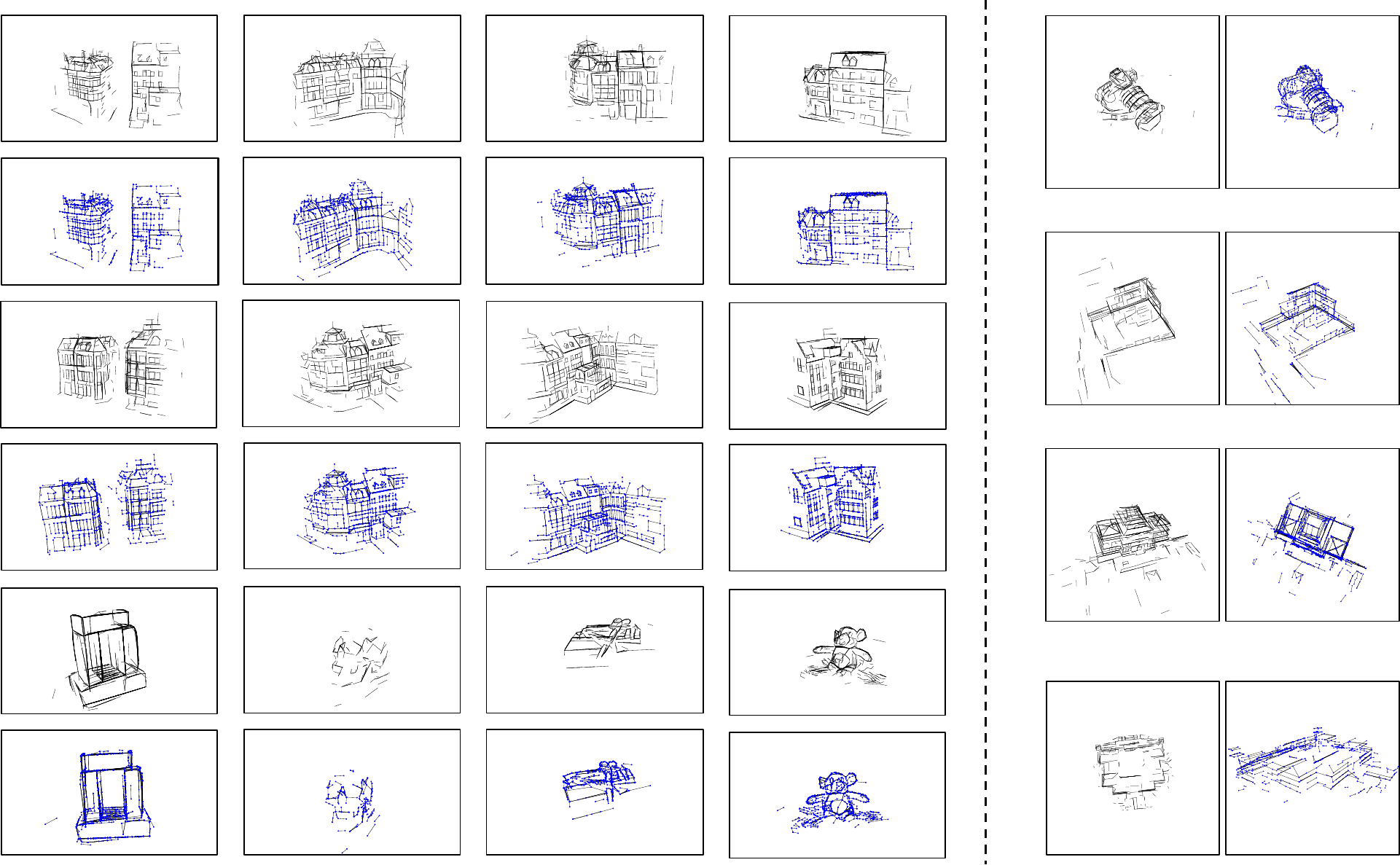}
    \caption{{\bf Visualization of 3D Wireframe Reconstruction} on the 12 scenes from the DTU dataset~\cite{DTU-AanaesJVTD16} and the 4 scenes from the BlendedMVS dataset~\cite{BMVS-dataset}. For each scene, we show its line segment view (by hiding the junctions) in black, and the wireframe view by coloring the junctions in \textcolor{blue}{blue}.
    For the comparison, please see our \href{https://youtu.be/qtBQYbOpVpc}{video}. }
    \label{fig:neat-show}
\end{figure*}

\begin{table*}[t]%
    \centering
    \caption{
    Evaluation Results on the DTU and BlendedMVS datasets for the reconstructed 3D wireframes. ACC-J and ACC-L are the evaluation for junctions and line segments. For Line3D++@HAWP, LiMAP and ELSR, all the endpoints of line segments are treated as junctions.
    }
    \resizebox{0.9\linewidth}{!}{
    \begin{tabular}{c|ccccc|cccc|cccc}
    \toprule 
              & \multicolumn{5}{c|}{\textbf{NEAT (Ours)}}   & \multicolumn{4}{c|}{LiMAP~\cite{limap}}& \multicolumn{4}{c}{Line3D++@HAWP} \\
         Scan & ACC-J $\downarrow$ & ACC-L & COMP-L $\downarrow$ & \#Lines $\uparrow$ & \#Junctions  & ACC-J $\downarrow$ & ACC-L & COMP-L $\downarrow$ & \#Lines $\uparrow$ & ACC-J $\downarrow$ & ACC-L $\downarrow$ & COMP-L $\downarrow$ & \#Lines $\uparrow$ \\\midrule
               \multicolumn{14}{c}{DTU Dataset~\cite{DTU-AanaesJVTD16}} \\\hline
         Avg. & \bf 0.7718 & \bf 0.8002 &    \bf 6.1064  &   624 & 503&  1.0944 &    0.8547  & 7.7756        &231      &0.9019      &0.8133&    8.5086  &249\\ \midrule
         16   & 0.8263 &	0.7879 &	5.4135 	&	729 & 554 & 1.0385 &	0.7898 	& 6.0420 	     &  335   & 0.7957& 	0.6992& 	6.9052 & 388 \\
         17   & 0.7754 &	0.6695 &	5.0498 	&	738 & 546 & 1.1015 &	0.8804 	& 5.8212 	     &  388   & 0.8816& 	0.7778& 	7.6257 & 395 \\
         18   & 0.6429 &	0.6868 &	5.3796 	&	701 & 596 & 0.9950 &	0.8253 	& 7.0154 	     &  287   & 0.7894& 	0.7528& 	7.7082 & 305 \\
         19   & 0.6989 &	0.6923 &	4.6529 	&	809 & 510 & 0.7689 &	0.7110 	& 7.9461 	     &  160   & 0.6815& 	0.7953& 	6.9776 & 330 \\
         21   & 0.9042 &	0.6923 &	4.6529 	&	809 & 571 & 1.1011 &	0.8884 	& 5.9821 	     &  319   & 0.9064& 	0.7953& 	6.9776 & 330 \\
         22   & 0.6343 &	0.6910 &	5.0871 	&	758 & 596 & 0.8998 &	0.7353 	& 6.8567 	     &  281   & 0.7494& 	0.7079& 	7.8014 & 328 \\
         23   & 0.5882 &	0.6193 &	5.5992 	&	771 & 597 & 1.0561 &	0.8293 	& 6.5078 	     &  377   & 0.8005& 	0.7356& 	8.2679 & 320 \\
         24   & 0.6386 &	0.5944 &	5.9104 	&	860 & 549 & 1.0314 &	0.8293 	& 6.5078 	     &  377   & 0.7940& 	0.6807& 	7.6886 & 366 \\ 
         37   & 1.4815 &	1.0856 &	7.5362 	&	420 & 405 & 1.2721 &	1.2352 	& 8.6413 	     &  120   & 1.1796& 	1.0287& 	10.2244 & 60 \\
         40   & 0.6298 &	1.0354 &	8.7825 	&	137 & 469 & 1.2108 &	0.8327 	& 9.9988       &  	41   & 0.8486& 	0.6877& 	10.1206 & 83 \\
         65   & 0.7212 &	1.0354 &	8.7825 	&	137 & 171 & 1.0469 &	0.5071 	& 11.1936      &   	7   & 1.1008& 	1.0697& 	11.1519 & 23 \\
         105  & 0.7204 &	1.0127 &	6.4296 	&	621 & 478 & 1.6108 &	1.1929 	& 10.7943      &  	90   & 1.2957& 	1.0286& 	10.6539 & 61 \\\midrule
         \multicolumn{14}{c}{BlendedMVS Dataset~\cite{BMVS-dataset}} \\\hline
         Avg. & \bf 0.1949 & \bf 0.1802 & \bf 6.4621  & 602 & 514 & 0.3712 & 0.3169 & 6.9415 & 313 &  0.3743 & 0.3545 & 6.8760 & 724\\ \midrule
         1    & 0.0365 & 0.0404 & 3.7253 & 653 & 565 & 0.0488 & 0.0651 & 5.0457 & 226 & 0.0682 & 0.0650 & 5.3625 & 691 \\
         2    & 0.1715 & 0.1585 & 8.2943 & 328 & 343 & 0.3478 & 0.2817 & 8.7663 & 195 & 0.4327 & 0.4174 & 8.8864 & 396 \\
         3    & 0.2564 & 0.2165 & 7.5600 & 931 & 664 & 0.3796 & 0.3162 & 7.5366 & 467 & 0.3795 & 0.3582 & 7.3192 & 931 \\
         4    & 0.3153 & 0.3055 & 6.2686 & 509 & 483 & 0.7086 & 0.6045 & 6.4174 & 365 & 0.6171 & 0.5774 & 5.9359 & 876 \\
    \bottomrule
    \end{tabular}
    }
    \vspace{-4mm}
    \label{tab:dtu-evaluation}
\end{table*}

\subsection{Baselines, Datasets and Evaluation Metrics}
We take the well-engineered Line3D++~\cite{HoferMB17-Line3D++} and the recently-proposed LiMAP~\cite{limap} as the baselines to make quantative and qualitative comparisons, all of which are mainly designed for line-based 3D reconstruction based on two-view line matching results.  Because our target is 3D wireframe reconstruction instead of 3D line segment reconstruction, for fair comparisons, we use HAWPv3~\cite{hawpv3} as the alternative for 2D detection in the use of Line3D++ and LiMAP. For those baselines, we use their official implementation for 3D line segments reconstruction.

\myparagraph{DTU~\cite{DTU-AanaesJVTD16} and BlendedMVS~\cite{BMVS-dataset} Datasets.} These two datasets were mainly designed for multiview stereo (MVS), but they are applicable to 3D wireframe reconstruction as they provided high-quality 3D point clouds as annotations. For our experiments, we run our method on {12 scenes} from DTU datasets and 4 scenes from BlendedMVS datasets. For the quantitative evaluation, we first convert the reconstructed wireframe model by NEAT (or the 3D line segment model by baselines) into the point cloud by sampling $32$ points on each line segment and computing the ACC metric to make comparisons. Because the reconstructed 3D wireframes (and line segments) are rather sparse than the dense surfaces, the COMP metric used for comparison would be less informative than ACC. Therefore, we additionally use the number of reconstructed 3D line segments and junctions as the reference of completeness.

\subsection{Main Comparisons}%

We compare our NEAT approach with three baselines on the scenes from DTU and BlendedMVS datasets, which include both the straight-line dominant scenes and some curve-based ones.
In Tab.~\ref{tab:dtu-evaluation}, we quantitatively report the ACCs for both 3D line segments and their junctions (or endpoints), as well as the number of geometric primitives. Compared to the baseline {\em Line3D++@HAWP} that takes the same 2D wireframes as input, our NEAT significantly outperforms it in all metrics, which indicates that NEAT is able to yield more accurate and complete 3D reconstruction results than L3D++ for HAWP inputs. %
Fig.~\ref{fig:neat-show} visualizes the reconstructed 3D wireframes for the evaluated scenes on the DTU and BlendedMVS datasets. %

\subsection{Ablation Studies}
In our ablation study, two scenes (\ie, DTU-24 and DTU-105) are used as representative cases to discuss our NEAT approach. In the first, we qualitatively show the NEAT lines (\ie, raw output of 3D line segments by querying the NEAT field), the initial reconstruction by binding the queried NEAT lines to global junctions, and the final reconstruction results by the visibility checking.
Then, we discuss our NEAT approach in the following two aspects: (1) the parameterization of NEAT Fields and (2) the view dependency issue for junction perceiving. %
For more ablation studies for the hyperparameter setting, especially for the number of global junctions, please refer to \cref{appx:finalization}.
\begin{figure}
    \centering
    \subfloat[5905 lines]{
    \fbox{
    \includegraphics[height=0.25\linewidth]{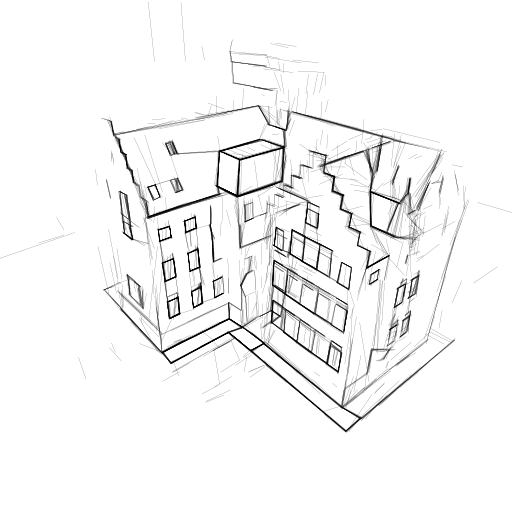}
    }
    }
    \subfloat[1399 lines]{
    \fbox{
    \includegraphics[height=0.25\linewidth]{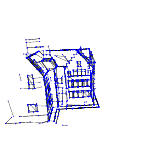}
    }
    }
    \subfloat[526 lines]{
    \fbox{
    \includegraphics[height=0.25\linewidth]{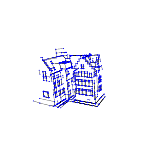}
    }
    }

    \subfloat[10394 lines]{
    \fbox{
    \includegraphics[height=0.25\linewidth]{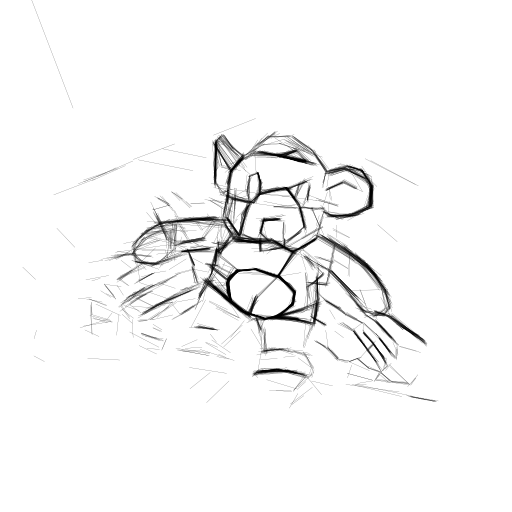}
    }
    }
    \subfloat[1102 lines]{
    \fbox{
    \includegraphics[height=0.25\linewidth]{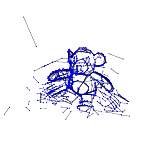}
    }
    }
    \subfloat[621 lines]{
    \fbox{
    \includegraphics[height=0.25\linewidth]{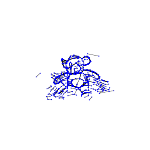}
    }
    }
    \caption{Left: NEAT lines (by coordinate MLP); Middle: initial wireframes (without visibility checking); Right: the final wireframes (with visibility checking) in the right.}
    \vspace{-4mm}
    \label{fig:illustration}
\end{figure}

\myparagraph{The Process of Wireframe Reconstruction.} Fig.~\ref{fig:illustration} shows the three components for wireframe reconstruction. In the first component, we query all possible 3D line segments from the optimized NEAT field. In the second component, the queried 3D line segments are binding to the global junctions. 
In the third step, by leveraging the non-linear optimization and a relaxed visibility checking, the unstable 3D line segments are removed from the initial wireframe models. Benefitting from the proposed novel mechanism of learning global 3D junctions, we largely simplified the way of removing duplicated and unreliable line segments without using either the known 3D points or the complicated line segment matching. 
\begin{table}[t]
    \centering
    \caption{Quantatively evaluation results for ablation studies on the DTU-24 and DTU-105 scenes.}
    \vspace{-4mm}
    \resizebox{\linewidth}{!}{
    \begin{tabular}{ccccccc}
    \toprule
                            & View Dir. & Clustering & ACC (J)$\downarrow$ & ACC (L)$\downarrow$ & \# Lines  & \# Junctions  \\\midrule
    \multirow{3}*{DTU-24}   &  No        & No          & 0.925   & 0.847   & 744      &  531 \\
                            &  Yes        & No          & 0.796   & 0.678  & 827      &  475 \\
                            &  Yes        & Yes          & {\bf 0.639}   & {\bf 0.594}  & {\bf 860}      &  {\bf 549} \\\midrule
    \multirow{3}*{DTU-105}  &  No        & No          & 0.822 & 1.209 & 607 &  499 \\
                            &  Yes        & No          & 0.749   & 1.154   & 557      &  408 \\
                            &  Yes        & Yes          & {\bf 0.720}   & {\bf 1.013}   & {\bf 621}      &  478 \\
    \bottomrule
    \end{tabular}
    }
    \label{tab:ablation-study}
\end{table}

\myparagraph{Parameterization of NEAT Fields.}
We found that the parameterization of NEAT Fields learning is playing in a vital role in the wireframe reconstruction. Even though our NEAT field aims at representing 3D line segments by the displacement vectors of the 3D points, the localization error in the detected 2D wireframes will possibly lead to some 3D line segments that cannot be well supported by high-quality 2D detection results missing. The information on view direction is a key factor to avoid this issue and yield more complete results. According to Tab.~\ref{tab:ablation-study}, the parameterization without the viewing directions will result in a coarser reconstruction with larger ACC errors for both 3D junctions and line segments while having fewer line segments although the number of global junctions is similar to the final model.

\myparagraph{Clustering in Junction Perceiving.}
The DBScan~\cite{dbscan} clustering is a key factor in accurately perceiving global junctions from the view-dependent coordinate MLP of the NEAT field. To verify this factor, we ablated the DBScan clustering to optimize MLPs on DTU-24 and DTU-105. Quantitatively reported in Tab.~\ref{tab:ablation-study}, although the parameterization of viewing direction largely reduced the ACC errors for both reconstructed junctions and line segments, the number of 3D junctions and line segments is also significantly reduced. When we enable the clustering during optimization, the lower-quality 3D local junctions (from the NEAT field) can be filtered, thus leading to an easy-to-optimize mode to yield more 3D junctions and line segments with fewer reconstruction errors.

\subsection{NEAT for 3D Gaussian Splatting}
\begin{figure}
    \centering
    \includegraphics[width=\linewidth]{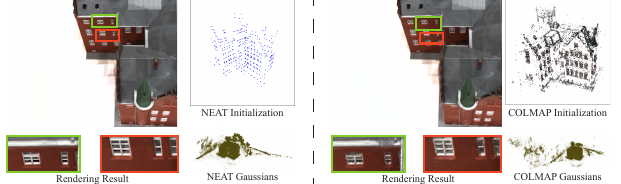}
    
    \caption{
    NEAT is applicable to 3D Gaussian Splatting framework to obtain more meaningful 3D Gaussian ellipsoids for better rendering results using 20 times fewer initial 3D points.
    }
    \vspace{-4mm}
    \label{fig:3dgs}
\end{figure}

Recently, 3D Gaussian Splatting~\cite{kerbl3Dgaussians} has become popular in neural rendering, owing to its computational efficiency and high-quality rendering. Our proposed NEAT method effectively represents 3D scenes using a limited number of junctions and line segments in wireframe format. We explored whether these reconstructed 3D junctions and line segments enhance novel view synthesis in 3D Gaussian Splatting~\cite{kerbl3Dgaussians} and found positive results. As demonstrated in \cref{fig:3dgs}, the final 3D Gaussian ellipsoids, optimized using different initialization (i.e., SfM Points and NEAT junctions), show that using only 549 points from the 3D junctions can yield more accurate geometry of Gaussian ellipsoids, thus improving rendering quality. Due to space constraints, further rendering experiments using NEAT's output are detailed in the \cref{appx:3dgs}.

\subsection{Failure Mode and Limitations}
\myparagraph{Volume Rendering of NEAT Fields.} Our method, based on VolSDF~\cite{YarivGKL21-VolSDF}, faces inherent difficulties in inside-out scenes for neural surface rendering, similar to recent studies~\cite{monosdf}. Overcoming these challenges, though possible with techniques like pre-trained monocular depth and normal maps~\cite{omnidata}, is beyond this paper's scope and reserved for future work. 

\begin{figure}
    \centering
    \subfloat[\scriptsize Groundtruth 3D Mesh]{
        \includegraphics[width=0.37\linewidth]{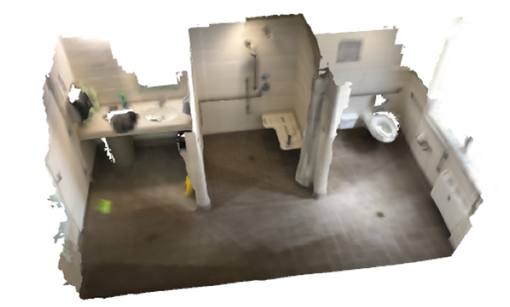}
    }
    \subfloat[\scriptsize Blur Images and Wireframes]{
    \includegraphics[width=0.37\linewidth]{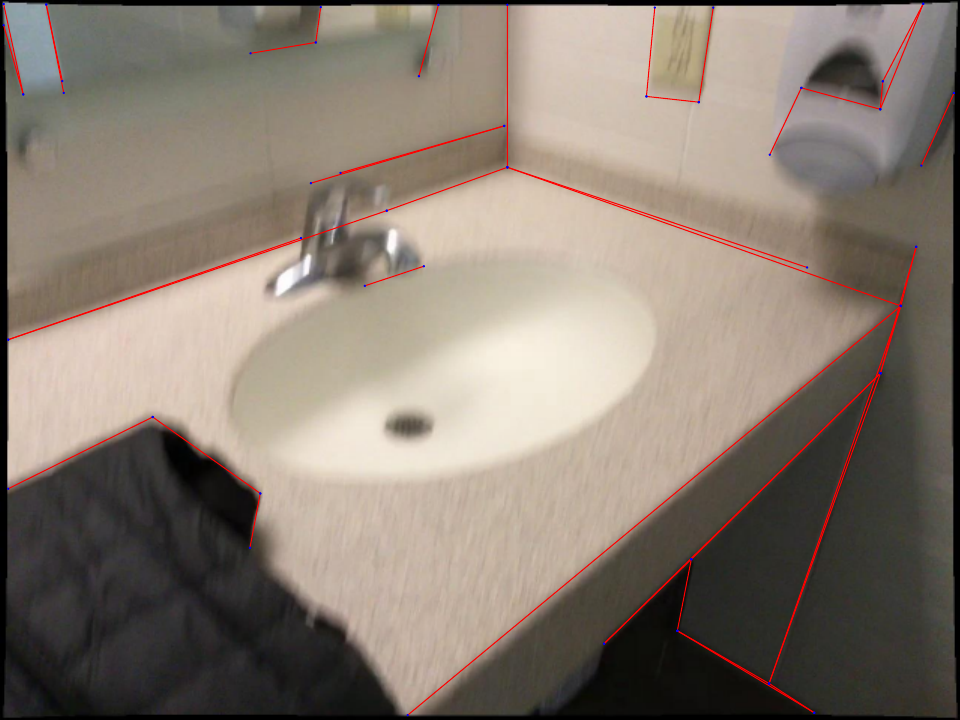}
    }\\
    
    \subfloat[\scriptsize Final Wireframe Model\label{fig:scan-wf}]{
    \fbox{
    \includegraphics[width=0.37\linewidth]{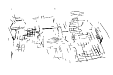}
    }
    }
    \subfloat[\scriptsize Global 3D Junctions\label{fig:scan-gjc}]{
    \fbox{
    \includegraphics[width=0.37\linewidth]{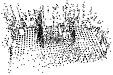}
    }
    }
    
    \caption{A Representative Failure Mode on ScanNet.}
    \vspace{-5mm}
    \label{fig:failure}
\end{figure}

\myparagraph{2D Detection Results are Critical.} Another critical issue is the quality of 2D wireframe detection. Failures in the HAWP model~\cite{hawpv3} directly impact our 3D wireframe reconstruction and parsing goals. Fig.\ref{fig:failure} illustrates a failure case from the ScanNet\cite{scannet} dataset, highlighting issues like motion blur affecting wireframe detection and leading to inaccuracies in 3D line segments. Despite these challenges, our global junctions (Fig.~\ref{fig:scan-gjc}) show potential in learning from blurry 2D wireframes, suggesting new insights into the relationship between junctions and line segments in wireframe representation.

\myparagraph{The Scalability Issue.} Our proposed method is currently limited by the predefined number of 3D global junctions (\eg 1024 junctions), which would be challenged in large-scale scenes that apparently contain much more 3D junctions. Though this limitation can be alleviated by leveraging a divide-and-conquer strategy like Block-NeRF~\cite{block-nerf}, the number of junctions should be scene-dependent and be automatically determined instead of being treated as a predefined hyperparameter in the future work.

%% file: sec/6_conclusion.tex
\section{Conclusion}%
This paper studied the problem of multi-view 3D wireframe parsing (reconstruction) to provide a novel viewpoint for compact 3D scene representation. Building on the basis of the volumetric rendering formulation, we propose a novel NEAT solution that simultaneously learns the coordinate MLPs for the implicit representation of the 3D line segments, and the global junction perceiving (GJP) to explicitly learn global junctions from the randomly-initialized latent arrays in a self-supervised paradigm. Based on new findings, we finally achieve our goal of computing a parsimonious 3D wireframe representation from 2D images and wireframes without considering any heuristic correspondence search for 2D wireframes. 
To our knowledge, we are the first to achieve multi-view 3D wireframe reconstruction with volumetric rendering. Our proposed novel junction perceiving module opens a door to characterize the scene geometry from 2D supervision in structured point-level 3D representation. 

{\small 
\myparagraph{Acknowledgment.}
N. Xue was partially supported by the NSFC under Grant 62101390. T. Wu was supported in part by NSF IIS-1909644. We would like to thank anonymous reviewers for their constructive suggestions. The views presented in this paper are those of the authors and should not be interpreted as representing any funding agencies.
}

%% file: appx/0_overview.tex
\clearpage
\setcounter{page}{1}
\maketitlesupplementary

The supplementary document is summarized as follows:
\begin{itemize}
    \item Appx.~\ref{appx:video} gives a summary of the supplementary video. %
    
    \item Appx.~\ref{appx:ray-sampling} elaborates on the technical details ({\em introduced in Sec. 3.2 of the main paper}) of NEAT optimization.
    
    \item Appx.~\ref{appx:finalization} supplies the details for the final step of distillation for 3D wireframe reconstruction ({\em introduced in Sec. 3.3} of the main paper).

    \item Appx.~\ref{appx:exp-abc} presents the additional experiments on the ABC dataset~\cite{abc-dataset} to discuss the performance given the ground-truth annotations of 3D wireframes. 

    \item Appx.~\ref{appx:3dgs} quantitatively reports the potential of NEAT for view synthesis with 3D Gaussian Splatting on the DTU dataset.

    \item Appx.~\ref{appx:misc} shows the miscellaneous stuff.

\end{itemize}

%% file: appx/1_video.tex
\appendix

\section{Video}\label{appx:video}
In our \href{https://youtu.be/qtBQYbOpVpc}{supplementary video}, we begin by demonstrating the core concepts of our research. Using a basic object from the ABC dataset as an illustrative example, we showcase the 3D line segments learned through the NEAT field, the functionality of the global junction perceiving module, and the construction of the final 3D wireframe model. Following this, the video highlights the learning of redundant 3D line segments and the optimization process for global junctions, using the DTU-24 dataset as a case study. The video concludes with qualitative evaluations on both the DTU and BlendedMVS datasets, providing visual support to the quantitative analyses of the main paper.

%% file: appx/2_raysampling.tex
\section{Optimization of NEAT}\label{appx:ray-sampling}

\begin{figure}[!h]
    \centering
    \subfloat[Input images]{
        \includegraphics[width=0.24\linewidth]{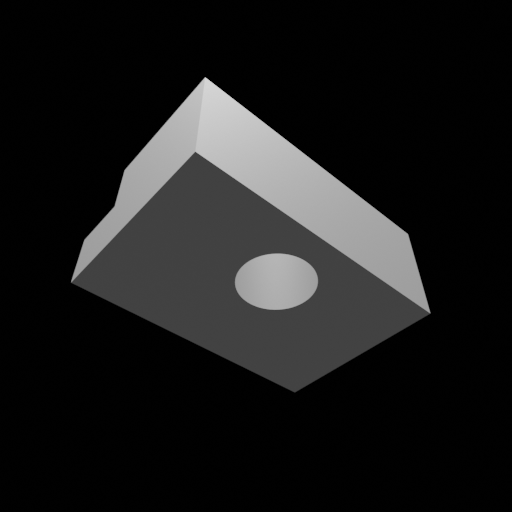}
        \includegraphics[width=0.24\linewidth]{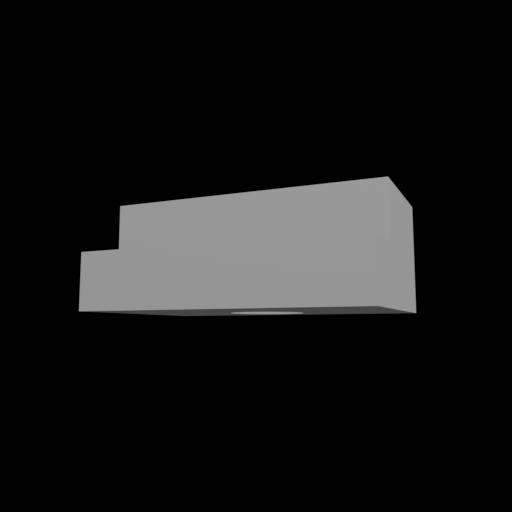}
        }
    \subfloat[2D Wireframes \& rendering pixels]{
        \includegraphics[width=0.24\linewidth]{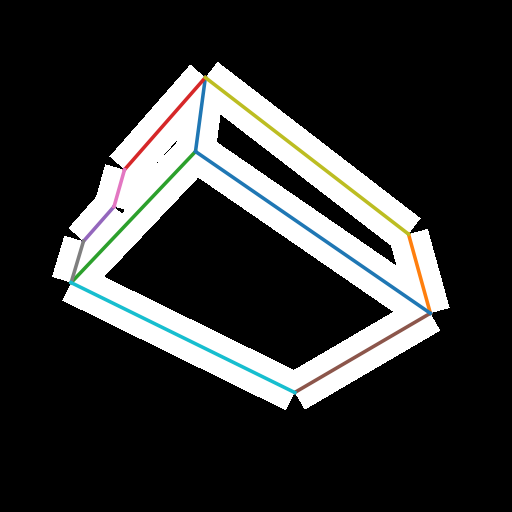}
        \includegraphics[width=0.24\linewidth]{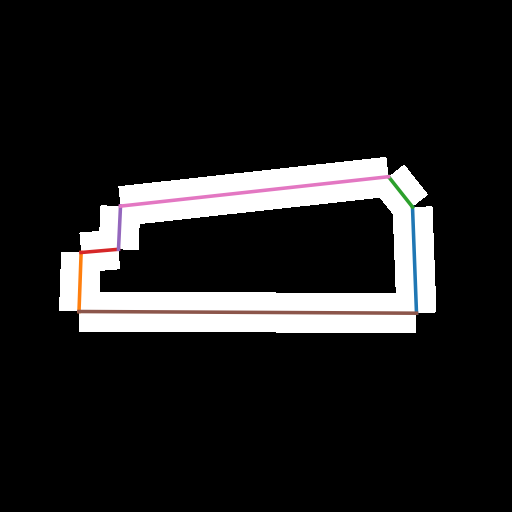}
    }
    \caption{A toy example on the ABC dataset~\cite{abc-dataset} for the foreground pixels defined by the detected 2D wireframes.}
    \label{fig:abc-toy-render-pix}
\end{figure}

\begin{figure}[!h]
    \centering
    \resizebox{\linewidth}{!}{
    \begin{tabular}{ccc}
    \fbox{
    \includegraphics[width=0.28\linewidth]{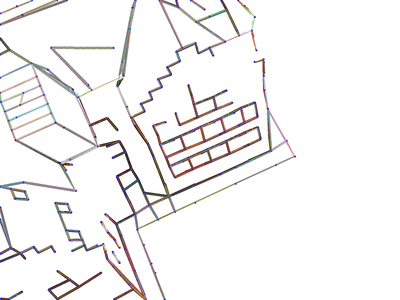}
    }&
    \fbox{
    \includegraphics[width=0.28\linewidth]{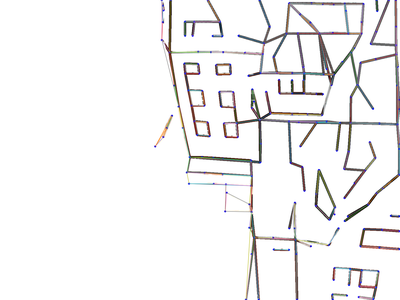}
    }&
    \fbox{
    \includegraphics[width=0.28\linewidth]{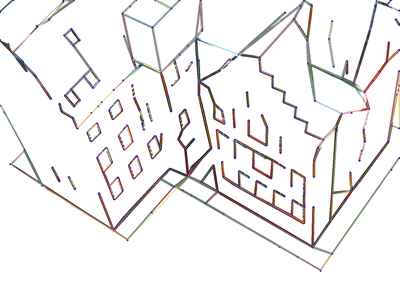}
    }\\
    {\tt MR: 90.88\%} & {\tt MR: 91.42\%} & {\tt MR: 89.32\%}
    \\ 
    \multicolumn{3}{c}{\scriptsize (a). Foreground Pixels defined by 2D wireframes ($\tau_d=5$)}\\
    
    \fbox{
    \includegraphics[width=0.28\linewidth]{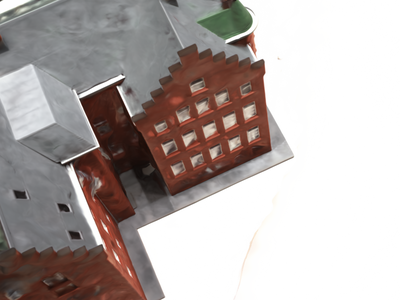}
    }&
    \fbox{
    \includegraphics[width=0.28\linewidth]{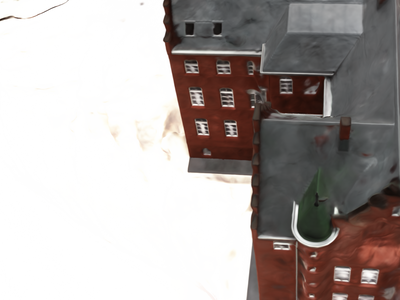}
    }&
    \fbox{
    \includegraphics[width=0.28\linewidth]{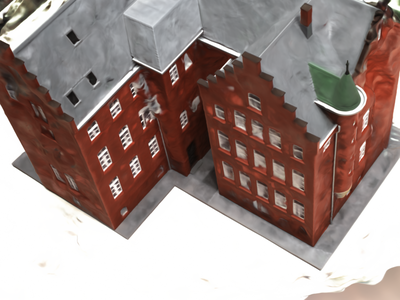}
    }\\
    {\footnotesize \tt PSNR: 25.46} & {\footnotesize \tt PSNR: 26.31} & {\footnotesize \tt PSNR: 21.37}
    \\
    \multicolumn{3}{c}{\scriptsize  (b). Rendered Images by NEAT}\\

    \fbox{
    \includegraphics[width=0.28\linewidth]{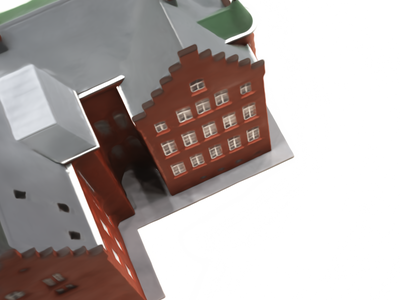}
    }&
    \fbox{
    \includegraphics[width=0.28\linewidth]{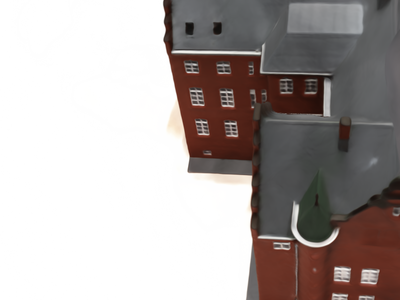}
    }&
    \fbox{
    \includegraphics[width=0.28\linewidth]{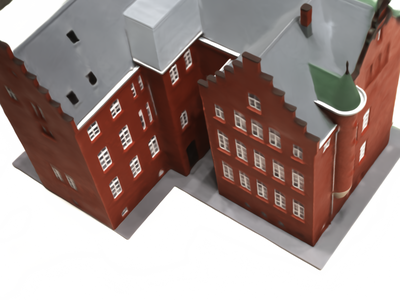}
    }\\
    {\footnotesize \tt PSNR: 27.40} & {\footnotesize \tt PSNR: 28.52} & {\footnotesize \tt PSNR: 26.62}
    \\
    \multicolumn{3}{c}{\scriptsize (c). Rendered Images by VolSDF~\cite{YarivGKL21-VolSDF}}\\

    \end{tabular}
    }
    \caption{A comparison for volumetric rendering learned from wireframe-related rays (pixels) \vs the vanilla ray sampling. In (a), we show the 2D line segments detected by HAWPv3~\cite{hawpv3} and the used foreground pixels in each view. ``{\tt MR}" denotes the mask ratio (the number of foreground pixels among all the pixels). In (b), we show the corresponding views rendered by NEAT that are learned by the foreground pixels in (a). In the bottom (c), we show the rendered images by VolSDF~\cite{YarivGKL21-VolSDF} as the reference. In (b) and (c), the PSNR values are marked at the bottom for each view.
    }
    \label{fig:masked-rend}
\end{figure}

\subsection{Details on Line Segment Rendering}
Our method renders 3D line segments based on the detected 2D wireframes in each 
view, distinguishing itself from conventional volume rendering approaches that 
utilize all pixels (rays) for rendering. As demonstrated in 
Fig.~\ref{fig:abc-toy-render-pix} with a toy example from the ABC dataset, 
only pixels with ``white" colors are engaged in the rendering process of 3D line 
segments. This technique is inspired by the attraction field representations
~\cite{afm-cvpr,afm-pami,hawp,hawpv3,deeplsd}, where the involved pixels are 
determined by the perpendicular distance between a point and a line segment. We 
set a threshold, $\tau_{\rm ray}$ (as mentioned in Sec.~\textcolor{red}{3.1} 
of our main paper), to differentiate the rendering pixels as foreground while 
disregarding the non-rendering pixels as background. Practically, 
$\tau_{\rm ray}$ is usually set to 5 for training/optimization, and reduced 
to 1 to minimize computational costs. We refer to this approach as 
{\em wireframe-driven ray sampling}.

\input{tables/distance-threshold}

To demonstrate the effectiveness of wireframe-driven ray sampling, we conducted 
a series of experiments on {\tt scene 24 from the DTU dataset}~\cite{DTU-AanaesJVTD16}. 
Fig.~\ref{fig:masked-rend} illustrates the feasibility of optimizing coordinate MLPs 
using this sampling technique. As depicted in Fig.~\ref{fig:masked-rend}(a), by 
masking over 80\% of the pixels (using a distance threshold of 5 pixels), we can 
still effectively optimize coordinate MLPs, leading to the reasonable outcomes 
shown in Fig.~\ref{fig:masked-rend}(b).

In addition to rendering results, we observed that increasing the distance threshold 
leads to a reduction in the number of line segments and junctions. As detailed in 
Tab.~\ref{tab:ablation-distance-threshold}, setting the distance threshold to 
$\tau_d = 20$ results in fewer 3D lines and junctions. Although the ACC errors 
are marginally reduced, there is an increase in completeness. Conversely, when 
the distance threshold $\tau_d$ is set to $1$, a performance degradation is 
noted across all metrics due to insufficient supervision signals.

\subsection{The Number of Global Junctions}
The number of global junctions is determined heuristically to encompass all potential 
3D junctions. Based on observations from both the DTU and BlendedMVS datasets, where 
the detected 2D line segments are in the hundreds, we set the estimated number of 
3D junctions to 1024. In \cref{tab:num-junctions}, we present experiments conducted 
on the DTU-24 scene with varying numbers of junctions, denoted as $N$, to assess 
performance differences. The results indicate that increasing the number of possible 
global 3D junctions to a larger value (e.g., $N=2048$) yields only a marginal increase 
in the count of learned 3D line segments and junctions in the final wireframe models. 
Conversely, a smaller $N$ tends to result in incomplete 3D wireframe models.

\begin{table}[!h]
    \centering
    \resizebox{\linewidth}{!}{
    \begin{tabular}{c|c|ccccc}
\toprule
$N$                 &  \# 2D Juncs.  & \# 3D Junctions & \# 3D Lines & ACC-J &  ACC-L & COMP-L\\\midrule
$1024$ (default)    &  \multirow{4}{*}{\makecell{212 (min)\\ 297 (max)\\ 258.2 (avg)}}                 & 549             & 860         & 0.639 & 0.549 & 5.910   \\\cmidrule{1-1}\cmidrule{3-7}
$N=128$   &                             & 99              & 93          & 0.422 & 0.440 & 8.541 \\
$N=512$   &                             & 397             & 641         & 0.526 & 0.574 & 6.302  \\
$N=2048$  &                             & 624             & 983         & 0.656 & 0.599 & 5.849           \\
\bottomrule
\end{tabular}
}
    \caption{The performance influence of wireframe reconstruction from different configuration of the number of 3D junctions during optimization.}
    \label{tab:num-junctions}
\end{table}

\subsection{Additional Implementation Details}\label{appx:details-neat}
\paragraph{Network Architecture.} The coordinate MLPs used in our NEAT approach are derived from VolSDF~\cite{YarivGKL21-VolSDF}, which contains three coordinate MLPs for SDF, the radiance field, and the NEAT field. For the MLP of SDF, it contains 8 layers with hidden layers of width 256 and a skip connection from the input to the 4th layer. The radiance field and the NEAT field share the same architecture with 4 layers with hidden layers of width 256 without skip connections. The proposed global junction perceiving (GJP) module contains two hidden layers and one decoding layer as described in the code snippets of Sec.~{\color{red} 1} in our main paper.

\paragraph{Hyperparameters.} The distance threshold $\tau_d$ about the foreground pixel (ray) generation is set to $5$ by default.%
For the number of global junctions (\ie, the size of the latent), we set it to $1024$ on the DTU and BlendedMVS datasets. When the scene scale is larger (\eg, a scene from ScanNet mentioned in Fig. 5 of the main paper), the number of global junctions is set to $2048$. For DBScan~\cite{dbscan}, we use the implementation from {\tt sklearn} package, set the epsilon (for the maximum distance between two samples) to 0.01 and the number of samples (in a neighborhood for a point to be considered as a core point) to 2. 

%% file: tables/distance-threshold.tex
\begin{table}
    \centering
    \caption{The influence of wireframe reconstruction results from different distance thresholds. The larger $\tau_d$ value is, the more line segments are involved in the optimization/learning.}
    \resizebox{\linewidth}{!}{
    \begin{tabular}{c|ccccc|cc}
    \toprule
                     & ACC-J$\downarrow$    & ACC-L$\downarrow$  & COMP-L$\downarrow$ & \#Lines & \#Junctions & MR & PSNR\\ \midrule
    $\tau_d =1$      & 0.853                 & 0.764             & 6.137 & 785     & 540 & 97.49\% & 17.79\\
    $\tau_d =5$      & 0.639                 & 0.594             & 5.910 & 860     & 528 & 89.70\% & 21.55\\
    $\tau_d =20$     & 0.578                 & 0.596 & 6.158             & 694     & 508 & 66.10\% & 24.68\\
    \bottomrule
    \end{tabular}
    }
    \label{tab:ablation-distance-threshold}
\end{table}

%% file: appx/impl-details.tex
\section{The Final Distillation Step of NEAT}\label{appx:finalization}
This section elaborates on the final distillation step required in our NEAT methodology 
for 3D wireframe reconstruction, with a particular focus on the extensive use of global 
junctions. We aim to provide a detailed insight into this crucial phase of the NEAT process.

\begin{figure}
    \centering
    \includegraphics[width=0.45\linewidth]{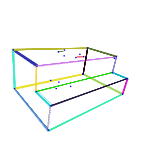}
    \includegraphics[width=0.45\linewidth]{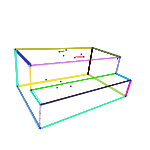}
    \caption{Two different views of the reconstruction of 3D wireframe on the toy scene of ABC dataset before the final distillation step.}
    \label{fig:finalize}
\end{figure}

To begin with, let us consider the challenge inherent in the junction-driven finalization 
of NEAT. As depicted in Fig.~\ref{fig:finalize}, using a toy ABC scene as an example, 
we observe that a considerable number of 3D line segments are rendered and aggregated 
across different views. Concurrently, 3D junctions are dynamically distilled from the 
NEAT fields. While a simple approach to combine these 3D junctions with the redundant 
3D line segments might seem viable, it is critical to address the potential misalignments 
between the junctions and line segments. To resolve this issue, we employ a least squares 
optimization combined with an SDF-based refinement scheme. This approach is designed 
to precisely adjust the position of 3D junctions, thereby ensuring an accurate and 
coherent reconstruction of the 3D wireframe.

\begin{table*}%
    \centering
    \caption{An Ablation study of the SDF-based 3D Junction Refinement on the DTU dataset for the reconstructed 3D wireframes. ACC-J and ACC-L are the evaluation for junctions and line segments.}
    \resizebox{0.9\linewidth}{!}{
    \begin{tabular}{c|cccc|cccc|cccc}
    \toprule 
              & \multicolumn{4}{c|}{NEAT (Final)}  & \multicolumn{4}{c}{NEAT (w/o Non-Linear Optimization)}  & \multicolumn{4}{c}{NEAT (w/o SDF-based Refinement)} \\
         Scan & ACC-J $\downarrow$ & ACC-L $\downarrow$ & \#Lines & \#Junctions  & ACC-J $\downarrow$ & ACC-L $\downarrow$ & \#Lines   & \#Junctions & ACC-J $\downarrow$ & ACC-L $\downarrow$ & \#Lines   & \#Junctions \\\midrule
         Avg. & 0.772 & 0.800 & 624.2 & 503.5 & 1.145 & 0.872 & 907.7 & 589.7 & 1.275 & 1.044 & 729.1 & 514.3 \\\hline
         16   & 0.826        & 0.788         & 729     & 554   & 0.834    & 0.829   & 852  & 566 & 1.190  & 1.045  & 751   & 570 \\
         17   & 0.775        & 0.670         & 738     & 546   & 0.982    & 0.765   & 991  & 651 & 1.047  & 0.836  & 753   & 557 \\
         18   & 0.643        & 0.687         & 701     & 596   & 0.930    & 0.759   & 993  & 689 & 1.040  & 0.927  & 821   & 609 \\
         19   & 0.699        & 0.692         & 809     & 510   & 0.956    & 0.703   & 994  & 656 & 1.051  & 0.863  & 714   & 518 \\
         21   & 0.904        & 0.692         & 809     & 571   & 0.960    & 0.725   & 981  & 654 & 1.119  & 0.848  & 816   & 581 \\
         22   & 0.634        & 0.691         & 758     & 596   & 0.896    & 0.748   & 939  & 684 & 0.976  & 0.897  & 769   & 603 \\
         23   & 0.588        & 0.619         & 771     & 597   & 0.840    & 0.703   & 933  & 670 & 0.926  & 0.821  & 774   & 602 \\
         24   & 0.639        & 0.594         & 860     & 549   & 0.818    & 0.620   & 1008  & 618 & 0.872  & 0.748  & 866   & 556 \\
         37   & 1.482        & 1.086         & 420     & 405   & 1.804    & 1.477   & 636  & 565 & 2.014  & 1.860  & 440   & 425  \\
         40   & 0.630        & 1.035         & 137     & 469   & 1.342    & 0.808   & 1672  & 591 & 1.382  & 0.983  & 1241  & 475 \\
         65   & 0.721        & 1.035         & 137     & 171   & 1.582    & 1.178   & 191  & 221 & 1.631  & 1.340  & 147   & 185  \\
         105  & 0.720        & 1.013         & 621     & 478   & 1.793    & 1.143   & 702  & 511 & 2.053  & 1.360  & 657   & 490 \\
    \bottomrule
    \end{tabular}
    }
    \vspace{-4mm}
    \label{tab:sdf-refinement-ablation}
\end{table*}
\subsection{Least Square Optimization}
To be convenient for readers, we copy Eq.~(9) in our main paper to \cref{eq:copy}, 
\begin{equation}\label{eq:copy}
   \mathcal{L}(J) = \sum_{(u,v)} \sum_{i=1}^{T_{u,v}} d_{\rm ang}(\mathbf{l}_{u,v}^0, \mathbf{l}_{u,v}^{i})^2 + d_{\rm perp}(\mathbf{l}_{u,v}^0, \mathbf{l}_{u,v}^{i})^2,
\end{equation}
which is the main objective function to adjust the junction positions according to the observation from the optimized/learned NEAT field. Here, we mathematically define the alignment cost between the junction-driven 3D line segments $\mathbf{l}_{u,v}^0 = (J_u,J_v)$ and its $i$-th NEAT-field observation $\mathbf{l}_{u,v}^{i} = (\mathbf{x}^i_u, \mathbf{x}^i_v)$ by the angular cost and the perpendicular cost as follow
\begin{equation}
\begin{split}
    d_{\rm ang}(\mathbf{l}_{u,v}^0,\mathbf{l}_{u,v}^{i}) & = 1 - |\langle \frac{J_u-J_v}{\left\|J_u-J_v\right\|}, \frac{\mathbf{x}_u^i-\mathbf{x}_v^i}{\left\|\mathbf{x}_u^i-\mathbf{x}_v^i\right\|}\rangle|, \\
    d_{\rm perp}(\mathbf{l}_{u,v}^0,\mathbf{l}_{u,v}^{i}) &= \left\|J_u - {\rm proj}(\mathbf{l}_{u,v}^i;J_u)\right\| \\
    & + \left\|J_v - {\rm proj}(\mathbf{l}_{u,v}^i;J_v)\right\|,
\end{split}
\end{equation}
where $\langle\cdot,\cdot\rangle$ is the inner product between two 3D vectors, and the function ${\rm proj}(\mathbf{l}_{u,v}^i;J_v)$ projects the point $J_v$ onto the infinite 3D line passing through the line segment $\mathbf{l}_{u,v}^i$. 
In Tab.~\ref{tab:sdf-refinement-ablation}, we report the performance changes by disabling the non-linear optimization on the DTU dataset, which will result in inferior 3D wireframes with larger ACC errors for both junctions and line segments. 

\subsection{SDF-based 3D Junction Refinement}
Following the non-linear optimization, we employ an SDF-based refinement scheme to 
further enhance the localization accuracy of junctions. Specifically, for an initial 
3D junction $J_i \in \mathbb{R}^3$ and an optimized SDF $d_{\Omega}(\cdot)$, we 
refine the location of $J_i$ using the following equation:
\begin{equation}
    J_i^{\rm refined} = J_i - d_{\Omega}(J_i) \cdot \nabla d_{\Omega}(J_i),
\end{equation}
where $\nabla d_{\Omega}$ represents the normal direction of the surface at the 
point $J_i$.

To assess the impact of this SDF-based refinement on junctions, we conducted an 
ablation study comparing 3D wireframe models with and without the SDF refinement. 
The results, presented in Tab.~\ref{tab:sdf-refinement-ablation}, clearly demonstrate 
the necessity of this refinement step for achieving significantly improved results.

\input{tables/eval-by-visibility}
\subsection{Visibility Checking}
As detailed in Sec.~3.3 of the main paper, we evaluate the reconstructed 3D line 
segments by projecting them onto 2D images from each view. This process involves 
computing both the angular and perpendicular distances between the projected 3D 
line segments and the detected 2D line segments. A 3D line segment is considered 
to be supported by a 2D detection if it aligns within an angular distance of 10 
degrees and a perpendicular distance of 5 pixels, with a minimum overlap ratio of 
50\%. This methodology allows us to determine the visibility of each 3D line 
segment and to filter out those that are invisible as false alarms.

In our standard approach, the visibility threshold for each line segment is set to 
$1$, aiming to achieve a more complete reconstruction. Moreover, we explore 
the impact of varying this visibility threshold from 1 to 4 on the DTU dataset. 
The findings, as summarized in \cref{tab:dtu-visibility-threshold}, indicate that 
increasing the visibility threshold results in an improvement in the ACC metric, while the COMP metric increases.

%% file: tables/eval-by-visibility.tex
\begin{table}
\caption{The performance change w.r.t. the visibility threshold on the DTU dataset.}\label{tab:dtu-visibility-threshold}
\resizebox{\linewidth}{!}{
\begin{tblr}{
cells={halign=c,valign=m},
column{1}={halign=r},
cell{2,6,10,14}{1}={r=4}{},
hline{1,18}={1.5pt},
hline{2}={1.0pt},
hline{6,10,14}={1.2pt},
vline{2,3,15},
}
Vis      & Metric      & 16    & 17    & 18    & 19    & 21    & 22    & 23    & 24    & 37    & 40    & 65    & 105 & Avg. \\
1 & ACC.$\downarrow$  & 0.788  & 0.670  & 0.687  & 0.692  & 0.692  & 0.691  & 0.619  & 0.594  & 1.086  & 1.035  & 1.035  & 1.013  & 0.800  \\
      & COMP.$\downarrow$ & 5.414  & 5.050  & 5.380  & 4.653  & 4.653  & 5.087  & 5.599  & 5.910  & 7.536  & 8.783  & 8.783  & 6.430  & 6.106  \\
      & Avg. Len. & 22.3  & 23.6  & 26.7  & 27.4  & 27.4  & 22.8  & 26.9  & 27.0  & 27.9  & 23.2  & 23.2  & 27.5  & 25.5  \\
      & \#Lines & 729.0  & 738.0  & 701.0  & 809.0  & 809.0  & 758.0  & 771.0  & 860.0  & 420.0  & 137.0  & 137.0  & 621.0  & 624.2  \\
2 & ACC.$\downarrow$  & 0.770  & 0.669  & 0.650  & 0.642  & 0.686  & 0.678  & 0.604  & 0.585  & 1.251  & 0.755  & 1.005  & 1.011  & 0.776  \\
      & COMP.$\downarrow$ & 5.493  & 5.067  & 5.043  & 5.562  & 4.742  & 5.208  & 5.670  & 6.032  & 7.517  & 7.027  & 9.131  & 6.643  & 6.095  \\
      & Avg. Len. & 22.3  & 23.6  & 24.4  & 27.0  & 27.6  & 22.8  & 26.9  & 27.1  & 27.4  & 49.8  & 22.8  & 27.0  & 27.4  \\
      & \#Lines & 711.0  & 729.0  & 789.0  & 667.0  & 784.0  & 737.0  & 756.0  & 840.0  & 391.0  & 1140.0  & 124.0  & 572.0  & 686.7  \\
3 & ACC.$\downarrow$  & 0.729  & 0.642  & 0.640  & 0.629  & 0.652  & 0.639  & 0.590  & 0.575  & 1.188  & 0.748  & 0.909  & 0.981  & 0.743  \\
      & COMP.$\downarrow$ & 5.551  & 5.095  & 5.117  & 5.742  & 4.843  & 5.357  & 5.720  & 6.113  & 7.473  & 7.182  & 9.076  & 6.785  & 6.171  \\
      & Avg. Len. & 22.5  & 23.7  & 24.5  & 27.2  & 27.8  & 22.7  & 26.9  & 27.2  & 27.7  & 49.9  & 22.8  & 26.9  & 27.5  \\
      & \#Lines   & 689.0  & 708.0  & 765.0  & 642.0  & 751.0  & 708.0  & 748.0  & 826.0  & 371.0  & 1091.0  & 112.0  & 544.0  & 662.9  \\
4 & ACC.$\downarrow$  & 0.704  & 0.619  & 0.623  & 0.617  & 0.607  & 0.632  & 0.583  & 0.556  & 1.118  & 0.735  & 0.891  & 0.945  & 0.719  \\
      & COMP.$\downarrow$ & 5.572  & 5.256  & 5.222  & 5.838  & 5.021  & 5.458  & 5.825  & 6.168  & 7.612  & 7.164  & 9.220  & 7.004  & 6.280  \\
      & Avg. Len. & 22.5  & 23.8  & 24.8  & 27.5  & 28.0  & 22.9  & 27.0  & 27.3  & 27.7  & 50.5  & 22.8  & 26.3  & 27.6  \\
      & \#Lines   & 672.0  & 679.0  & 737.0  & 617.0  & 723.0  & 683.0  & 721.0  & 806.0  & 347.0  & 1052.0  & 97.0  & 501.0  & 636.3  \\
\end{tblr}
}
\end{table}

%% file: appx/4_additional_exp_abc.tex
\section{Experiments on the ABC Dataset}\label{appx:exp-abc}
\begin{figure}
    \centering
    \resizebox{0.9\linewidth}{!}{
    \begin{tabular}{lc}
    \rotatebox{90}{~~~~~Images}&
    \includegraphics[width=0.23\linewidth]{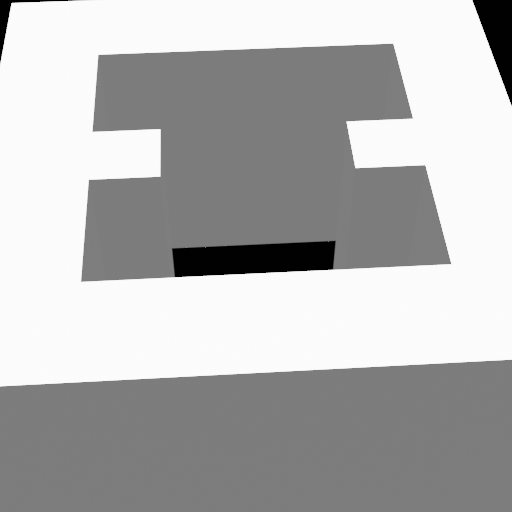}
    \includegraphics[width=0.23\linewidth]{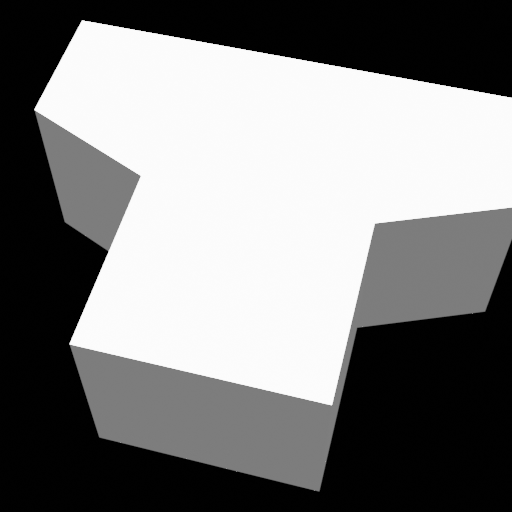}
    \includegraphics[width=0.23\linewidth]{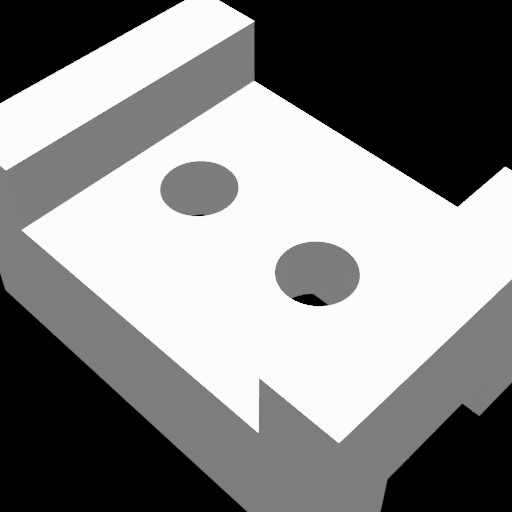}
    \includegraphics[width=0.23\linewidth]{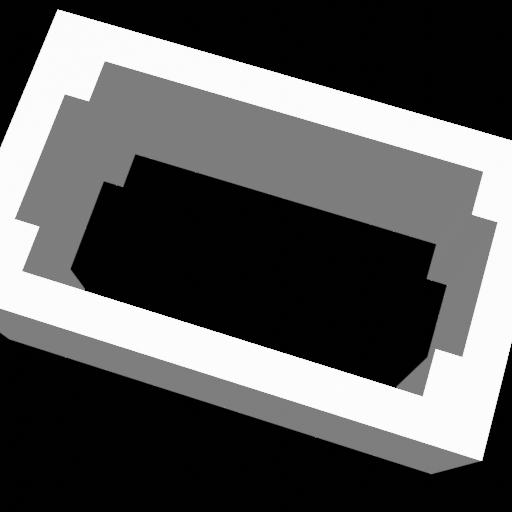}\\\hline\\
    \multirow{2}*{\rotatebox{90}{\small NEAT (Ours)~~~~~~}}&
    \fbox{\includegraphics[width=0.23\linewidth]{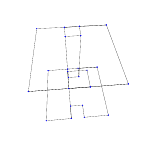}}
    \fbox{\includegraphics[width=0.23\linewidth]{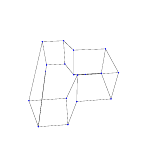}}
    \fbox{\includegraphics[width=0.23\linewidth]{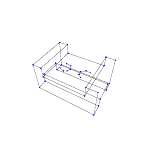}}
    \fbox{\includegraphics[width=0.23\linewidth]{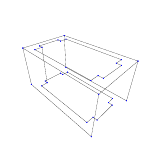}}\\
    & 
    \fbox{\includegraphics[width=0.23\linewidth]{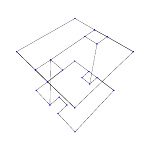}}
    \fbox{\includegraphics[width=0.23\linewidth]{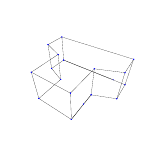}}
    \fbox{\includegraphics[width=0.23\linewidth]{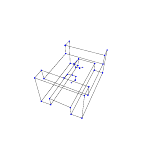}}
    \fbox{\includegraphics[width=0.23\linewidth]{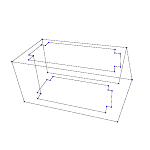}}\\
     & 
    \fbox{\includegraphics[width=0.23\linewidth]{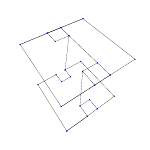}}
    \fbox{\includegraphics[width=0.23\linewidth]{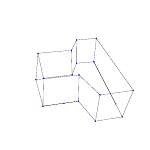}}
    \fbox{\includegraphics[width=0.23\linewidth]{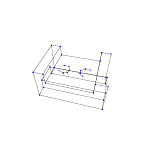}}
    \fbox{\includegraphics[width=0.23\linewidth]{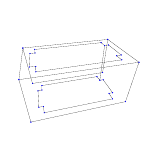}}\\\hline\\
    \rotatebox{90}{\small Ideal Baseline}&
    \fbox{\includegraphics[width=0.23\linewidth]{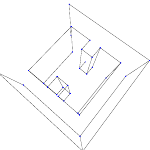}}
    \fbox{\includegraphics[width=0.23\linewidth]{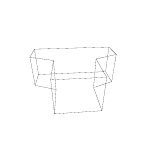}}
    \fbox{\includegraphics[width=0.23\linewidth]{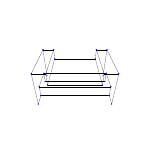}}
    \fbox{\includegraphics[width=0.23\linewidth]{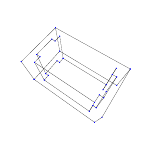}}
    \end{tabular}
    }
    \caption{Qualitative Comparisons on ABC objects.}
    \vspace{-3mm}
    \label{fig:abc-show}
\end{figure}

Because the 3D wireframe annotations are very difficult to obtain for real scene images, to better discuss the problem of 3D wireframe reconstruction and analyze our proposed NEAT approach, we conduct experiments on objects from ABC Datasets as it provides 3D wireframe annotations. 

\paragraph{Data Preparation.}
We use Blender~\cite{blender} to render 4 objects from the ABC dataset. The object IDs are mentioned in Tab.~\ref{tab:abc-table}. For each object, we first resize it into a unit cube by dividing the size of the longest side and then moving it to the origin center. Then, we randomly generate 100 camera locations, each of which is distant from the origin by $\sqrt{1.5^2+1.5^2} \approx 2.1213$ units. The setting of the distance, $\sqrt{1.5^2+1.5^2}$, is from our early-stage development for the rendering, in which we set a camera at $(0,1.5,1.5)$ location. By setting the cameras to look at the origin $(0,0,0)$, we obtain 100 camera poses.   
Considering the fact that the ABC dataset is relatively simple, we set the focal length to $60.00$mm to ensure the object is slightly occluded for rendering images. The sensor width and height of the camera in Blender are all set to $32$mm. The ground truth annotations of the 3D wireframe are from the corresponding \texttt{STEP} files.
For the simplicity of evaluation, we only keep the straight-line structures and ignore the curvature structures to obtain the ground truth annotations. The rendered images are with the size of $512\times 512$.

\paragraph{Baseline Configuration.} Fig.~\ref{fig:abc-show} illustrates the rendered input images for the used four objects. Because the rendered images are textureless and with planar objects, the dependency of those baselines on the correspondence-based sparse reconstruction by SfM systems~\cite{colmapCVPR16} is hardly satisfied to produce reliable line segment matches for 3D line reconstruction. Accordingly, we set up an ideal baseline instead of using Line3D++~\cite{HoferMB17-Line3D++} and LiMAP~\cite{limap} for comparison. Specifically, we first detect the 2D wireframes for the rendered input images and then project the junctions and line segments of the ground-truth 3D wireframe models onto the 2D image plane. For the 2D junctions, if a projected ground-truth junction can be supported by a detected one within $5$ pixels in any view, we keep the ground-truth junction as the reconstructed one in the ideal case. For the 2D line segments, we compute the minimal value for the distance of the two endpoints of a detected line segment to check if it can support a ground-truth 3D line. The threshold is also set to $5$ pixels. Then, we count the number of reconstructed 3D line segments and junctions in such an ideal case.

\paragraph{Evaluation Metrics.} For our method, we compute the precision and recall for the reconstructed 3D junctions and line segments under the given thresholds. Because the objects (and the ground-truth wireframes) are normalized in a unit cube, we set the matching thresholds to $\{0.01,0.02,0.05\}$ for evaluation. For the matching distance of line segments, we use the maximal value of the matching distance between two endpoints to identify if a line segment is successfully reconstructed under the specific distance threshold. For the ideal baseline, we report the number of ground-truth primitives (junctions or line segments), the number of reconstructed primitives, and the reconstruction rate.

\paragraph{Results and Discussion.} Tab.~\ref{tab:abc-table} quantitatively summarizes the evaluation results and the statistics on the used scenes. As it is reported, our NEAT approach could accurately reconstruct the wireframes from posed multiview images. The main performance bottleneck of our method comes from the 2D detection results. As shown in the ideal baseline, by projecting the 3D junctions and line segments into the image planes to obtain the ideal 2D detection results, the 2D detection results by HAWPv3~\cite{hawpv3} did not perfectly hit all ground-truth annotations. Furthermore, suppose we use the hit (localization error is less than 5 pixels) ground truth for 3D wireframe reconstruction, there is a chance to miss some 3D junctions and more 3D line segments. In this sense, given a relaxed threshold of the reconstruction error for precision and recall computation, our NEAT approach is comparable with the performance of the ideal solution. 
For the first object (ID 4981), because of the severe self-occlusion, some line segments are not successfully reconstructed for both the ideal baseline and our approach.
For object 17078, our NEAT approach reconstructed some parts of the two circles that are excluded from the ground truth, which leads to a relatively low precision rate.
Fig.~\ref{fig:abc-show} also supported our results.

\begin{table}[t]
    \centering
    \resizebox{0.9\linewidth}{!}{
    \begin{tabular}{cc|cccccc|ccc}
    \toprule
                            & & \multicolumn{6}{c|}{Evaluation Results} & \multicolumn{3}{c}{Ideal Baseline}\\
     ID                     & & $P_{0.01}$ & $P_{0.02}$ &$P_{0.05}$ & $R_{0.01}$ & $R_{0.02}$ & $R_{0.05}$ & \#GT   & \# Reconstructed & Recon. Rate\\\midrule
     \multirow{2}*{4981}    & J & 0.706    & 0.765      & 0.882     & 0.750      & 0.812      & 0.938 & 32 & 28 & 0.875\\
                            & L & 0.758 & 0.758 & 0.758 & 0.521 & 0.521 & 0.521 & 48 & 41 & 0.854\\\midrule
     \multirow{2}*{13166}   & J & 0.889    & 0.889      & 0.889     & 1.000      & 1.000      & 1.000 & 16 & 16 & 1.000\\
                            & L & 1.000    & 1.000      & 1.000     & 1.000      & 1.000      & 1.000 & 24 & 24 & 1.000\\\midrule
     \multirow{2}*{17078}   & J & 0.400    & 0.629      & 0.686     & 0.583      & 0.917      & 1.000 & 24 & 23 & 0.958\\
                            & L & 0.408    & 0.653      & 0.714     & 0.556      & 0.889      & 0.972 & 36 & 32 & 0.889\\\midrule
     \multirow{2}*{19674}   & J & 0.969    & 1.000      & 1.000     & 0.969      & 1.000      & 1.000 & 32 & 32 & 1.000\\
                            & L & 0.969 & 1.000 & 1.000 & 0.969 & 1.000 & 1.000  & 48 & 40 & 0.833\\    
     \bottomrule
    \end{tabular}
    }
    \caption{Evaluation Results and some Statistics on ABC objects. In each object, we evaluate the precision and recall rates for junctions (J) and line segments (L). For the ideal baseline, we count the number of ground-truth primitives, the number of reconstructed 3D primitives, and the reconstruction rate in the ideal baseline.}
    \vspace{-3mm}
    \label{tab:abc-table}
\end{table}

%% file: appx/gaussian-rendering.tex
\section{3D Gaussians with NEAT Junctions}\label{appx:3dgs}
In this section, we extend the application of our NEAT framework to 3D Gaussian 
Splatting, as proposed by Kerbl et al.~\cite{kerbl3Dgaussians}, by substituting 
the initial point cloud derived from Structure-from-Motion (SfM) with the junctions 
identified by NEAT. This experiment is designed to showcase the efficacy of NEAT 
junctions as a compact initialization method for 3D Gaussian Splatting. 
Using only a few hundred points, our NEAT junctions demonstrate an enhanced fitting 
ability on the DTU dataset, as evidenced by improved metrics in both Peak Signal-to-Noise 
Ratio (PSNR) and Structural Similarity Index (SSIM).

The experimental results on 12 scenes from the DTU dataset are detailed in 
\cref{tab:gaussian}. It is observed that by initializing the 3D Gaussians with 
NEAT junctions, there is a notable improvement in performance: PSNR increases 
by 0.38 dB and SSIM improves by 0.0003 points. This finding underscores the 
effectiveness of NEAT junctions in providing a more precise and compact starting 
point for 3D Gaussian Splatting.

\input{tables/gaussian}

%% file: tables/gaussian.tex
\begin{table}[]
\newcommand{\gain}[1]{\,(\textcolor{cyan}{#1})}
\newcommand{\drop}[1]{\,(\textcolor{olive}{#1})}
\centering
\SetTblrInner{rowsep=1.0pt}      %
\SetTblrInner{colsep=6pt}      %
\caption{Quantitative comparison between the NEAT junctions and SfM points for the initialization of 3D Gaussian Splatting on the DTU dataset.}\label{tab:gaussian}
\resizebox{\linewidth}{!}{
    \begin{tblr}{
        cells={halign=c,valign=m},   %
        column{1}={halign=l},        %
        cell{1}{1}={r=2}{},          %
        cell{1}{2,7}={c=5}{},   %
        hline{1,3,15,16} = {1-11}{},
        hline{1,16} = {1.5pt},
        hline{2} = {2-11}{},
        vline{2,4,7,9} = {1-15}{},
        vline{4,9} = {dashed}
    }
    Scene ID & NEAT Junctions & & & & & SfM Points (by COLMAP~\cite{colmapCVPR16}) & & & & \\
& PSNR $\uparrow$& SSIM $\uparrow$ & \makecell{\#Points \\(init)} & \makecell{\#Points\\ (7k)} & \makecell{\#Points\\ (30k)}&  PSNR $\uparrow$ & SSIM $\uparrow$ & \makecell{\#Points \\(init)} & \makecell{\#Points\\ (7k)} & \makecell{\#Points\\ (30k)}\\
DTU-16    & \textbf{28.7}\gain{+0.7}  & \textbf{0.889}\gain{+0.006} & 554   & 603k  & 1,496k & 28.0  & 0.883  & 22k   & 558k  & 1,048k \\
DTU-17    & \textbf{29.2}\gain{+0.5} & \textbf{0.898}\gain{+0.005} & 546   & 903k  & 2,279k & 28.7  & 0.893  & 24k   & 893k  & 1,305k \\
DTU-18    & \textbf{29.3}\gain{+0.4} & \textbf{0.901}\gain{+0.004} & 596   & 629k  & 1,234k & 28.9  & 0.897  & 18k   & 581k & 1,078k \\
DTU-19    & \textbf{29.6}\gain{+0.4} & 0.893\gain{-0.001} & 510   & 475k  & 1,140k & 29.2  & \textbf{0.894 } & 19k   & 561k  & 756k \\
DTU-21    & \textbf{28.7}\gain{+0.2} & \textbf{0.898}\gain{+0.004} & 571   & 725k  & 1,657k & 28.5  & 0.894  & 19k   & 698k  & 1,528k \\
DTU-22    & \textbf{29.1}\gain{+0.2} & \textbf{0.892}\gain{+0.005} & 596   & 641k  & 1,455k & 28.9  & 0.887  & 21k   & 615k  & 1,113k \\
DTU-23    & \textbf{28.4}\gain{+0.4} & \textbf{0.886}\gain{+0.006} & 597   & 974k  & 2,243k & 28.0  & 0.880  & 25k   & 850k  & 1,667k \\
DTU-24    & \textbf{31.1}\gain{+0.9} & \textbf{0.909}\gain{+0.008} & 549   & 587k  & 1,181k & 30.2  & 0.901  & 13k   & 528k  & 852k \\
DTU-37    & \textbf{28.2}\gain{+0.5} & \textbf{0.875}\gain{+0.000} & 405   & 420k  & 1,180k & 27.7  & \textbf{0.875 } & 27k   & 409k  & 713k \\
DTU-40    & \textbf{30.6}\gain{+0.2} & \textbf{0.862}\gain{+0.002} & 422   & 520k  & 1,403k & 30.4  & 0.860  & 32k   & 515k  & 1,070k \\
DTU-65    & \textbf{32.4}\gain{+0.2} & 0.855\drop{-0.001} & 171   & 139k  & 294k  & 32.2  & \textbf{0.856 } & 11k   & 150k  & 208k \\
DTU-105   & 30.8\drop{-0.1}  & 0.852\drop{-0.001} & 478   & 165k  & 238k  & \textbf{30.9 } & \textbf{0.853 } & 23k   & 169k  & 216k \\
\quad Avg.      & \textbf{29.68 }\gain{+0.38} & \textbf{0.884 }\gain{+0.003} & 499.58  & 565k  & 1,317k & 29.30  & 0.881  & 21k   & 544k  & 963k \\
    \end{tblr}}
\end{table}

%% file: appx/5_misc.tex
\section{Miscellaneous}\label{appx:misc}

\subsection{Evaluation Metrics}
\paragraph{The Definition of ACC and COMP Metrics.} We follow the official evaluation protocol of the DTU dataset~\cite{DTU-AanaesJVTD16} to compute the reconstruction accuracy (ACC) and completeness (COMP), which is defined to 
\begin{equation}
    {\rm ACC} = \underset{\substack{\mathbf{p}\in P}}{\mathrm{mean}} \left(\min_{\mathbf{p}^*\in P^*}\left\|\mathbf{p} - \mathbf{p}^*\right\|\right),
\end{equation}
and 
\begin{equation}
    {\rm COMP} = \underset{\substack{\mathbf{p^*}\in P^*}}{\mathrm{mean}} \left(\min_{\mathbf{p}\in P}\left\|\mathbf{p} - \mathbf{p}^*\right\|\right),
\end{equation}
where $P$ and $P^*$ are the point clouds sampled from the predictions and the ground truth mesh. %

\subsection{Information of Used BlendedMVS Scenes}
The scene IDs and their MD5 code of the BlendedMVS scenes are:
\begin{itemize}
    \item Scene-01:  \texttt{5c34300a73a8df509add216d}
    \item Scene-02:  \texttt{5b6e716d67b396324c2d77cb}
    \item Scene-03:  \texttt{5b6eff8b67b396324c5b2672}
    \item Scene-04:  \texttt{5af28cea59bc705737003253}
\end{itemize}